\documentclass[11pt]{article}
\usepackage[letterpaper, left=1truein, right=1truein, top = 1truein, bottom = 1truein]{geometry}
\usepackage{amssymb} 
\usepackage{amsmath}
\usepackage{graphicx}
\usepackage{natbib}
\usepackage{authblk}
\usepackage{lipsum} 
\usepackage{fancyhdr}
\newcommand{\citealtt}[1]{\citeauthor{#1},\ \citeyear{#1}}
\newcommand{\One}{I}
\newcommand{\disT}{\textstyle}
\newcommand{\disS}{\displaystyle}
\newcommand{\FF}{{\cal F}}
\newcommand{\sVec}{\vec{s}}

\newcommand{\sVecPrime}{\vec{s}^{\,\prime}}
\newcommand{\zVec}{\vec{z}}

\newcommand{\yVec}{\vec{y}}
\newcommand{\yVecN}{\vec{y}^{\,(n)}}
\newcommand{\piVec}{\vec{\pi}}
\newcommand{\Wt}{\tilde{W}}
\newcommand{\SCal}{\mathcal{S}}
\newcommand{\kappaVec}{\vec{\kappa}}
\newcommand{\muVec}{\vec{\mu}}
\newcommand{\kappaVecN}{\kappaVec^{(n)}}

\newcommand{\E}[1]{\big\langle{}#1\big\rangle}
\newcommand{\HPrime}{H^{\prime}}
\newcommand{\KK}{{\cal K}}
\newcommand{\KKn}{\KK_n}
\newcommand{\RRR}{\mathbb{R}}

\newcommand{\ThetaOld}{\Theta^{\mathrm{old}}}
\newcommand{\ThetaOLD}{\Theta^{\mathrm{old}}}
\newcommand{\NGauss}{{\cal N}}
\newcommand{\Bernoulli}{\mathcal{B}}
\newcommand{\Scal}{{\cal S}}
\newcommand{\sig}{\sigma}
\newcommand{\dz}{\mathrm{d}\zVec}
\newcommand{\TT}{\mathrm{T}}
\newcommand{\refp}[1]{(\ref{#1})}
\newcommand{\ssb}{\hspace{-2mm}}

\newcommand{\Sigmad}{\Sigma}
\newcommand{\Sigmah}{\Psi}
\newcommand{\BigO}{\mathcal{O}}
\newcommand{\trace}{\mathrm{Tr}}
\newcommand{\Qn}{Q^{(n)}}
\newcommand{\DKL}{D_{KL}}

\title{A Truncated EM Approach for\\ Spike-and-Slab Sparse Coding}

\author{Abdul-Saboor Sheikh$^1$,
~ Jacquelyn A.~Shelton$^1$ 
~ and ~ J\"org L\"ucke$^{2}$\\
\texttt{\small \{sheikh, shelton\}@tu-berlin.de, joerg.luecke@uni-oldenburg.de}\\ \vspace{2mm}
$^1$Faculty of Electrical Engineering and Computer Science\\
Technical University of Berlin, Marchstr.\ 23, 10587 Berlin, Germany\\ \vspace{2mm}
$^2$Cluster of Excellence Hearing4all and Faculty VI\\
University of Oldenburg, 26115 Oldenburg, Germany
}

\date{}

\begin{document}

\maketitle

\begin{abstract}
We study inference and learning based on a sparse coding model
with `spike-and-slab' prior. As in standard sparse coding, the
  model used assumes independent latent sources that linearly combine to
  generate data points. However, instead of using a standard sparse 
  prior such as a Laplace distribution, we study the application of a more flexible `spike-and-slab'
  distribution which models the absence or presence of a source's
  contribution independently of its strength if it
  contributes.
  We investigate two approaches to optimize the parameters of
  spike-and-slab sparse coding: a novel truncated EM approach and, for comparison, an approach based on
  standard factored variational distributions. The truncated approach can be regarded as a variational approach
with truncated posteriors as variational distributions.
  In applications to source separation we find that both approaches
  improve the state-of-the-art in a number of standard benchmarks,
  which argues for the use of `spike-and-slab' priors for the corresponding
data domains.
  Furthermore, we find that the truncated EM approach improves
  on the standard factored approach in source separation tasks---which
  hints to biases introduced by assuming posterior independence in the
  factored variational approach.
  Likewise, on a standard benchmark for image denoising, we find that
  the truncated EM approach improves on
  the factored variational approach.  While the performance of the factored approach
  saturates with increasing numbers of hidden dimensions, the performance
  of the truncated approach improves the state-of-the-art for higher noise levels.
\end{abstract}


\section{Introduction}
\label{SecIntro}
Much attention has recently been devoted to studying sparse coding
models with `spike-and-slab' distribution as a prior over the latent variables 
\citep{GoodfellowEtAl2013,MohamedEtAl2012,LuckeSheikh2012,TitsiasLazaro2011,CarbonettoStephen2011,KnowlesGhahramani2010,YoshidaEtAl2010}. 
In general, a `spike-and-slab' distribution is comprised of a binary (the
`spike') and a continuous (the `slab') part. The distribution generates a 
random variable by multiplying together the two parts such that the resulting 
value is either exactly zero (due to the binary random variable being zero) or it is 
a value drawn from a distribution governing the 
continuous part. In sparse coding models, employing spike-and-slab as a prior
allows for modeling the presence or absence of
latents independently of their contributions in generating an observation.
For example, piano keys (as latent variables) are either pressed or
not (binary part), and if they are pressed, they result in sounds with
different intensities (continuous part). The sounds
generated by a piano are also sparse in the sense that of all keys
only a relatively small number is pressed on average.

Spike-and-slab distributions can flexibly model an array of sparse
distributions, making them desirable for many types of data.
Algorithms based on spike-and-slab distributions have 
successfully been used, e.g., for deep learning and transfer learning \citep{GoodfellowEtAl2013}, 
regression~\citep{West2003,CarvalhoEtAl2008,CarbonettoStephen2011,TitsiasLazaro2011}, or denoising 
\citep{ZhouetAl2009,TitsiasLazaro2011}, and often
represent the state-of-the-art on given benchmarks \citep[compare][]{TitsiasLazaro2011,GoodfellowEtAl2013}.

\begin{figure}[t]
\begin{center}
\begin{minipage}[b]{5cm}
\includegraphics[width=1.\textwidth]{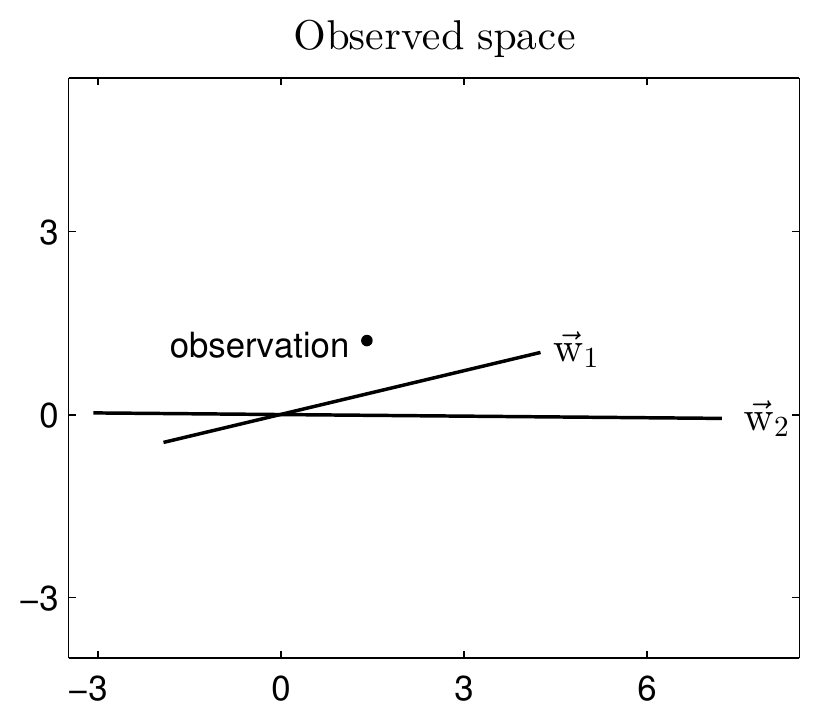}
\end{minipage}
\raisebox{2.05cm}{
\begin{minipage}{5cm}
\includegraphics[width=1.\textwidth]{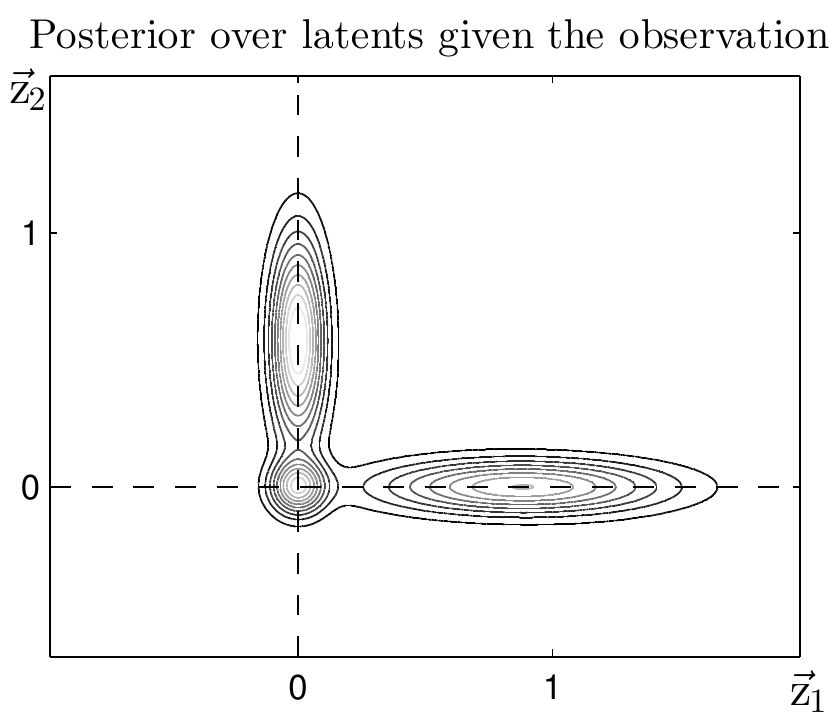}
\end{minipage} }
\begin{minipage}[b]{5cm}
\includegraphics[width=1.\textwidth]{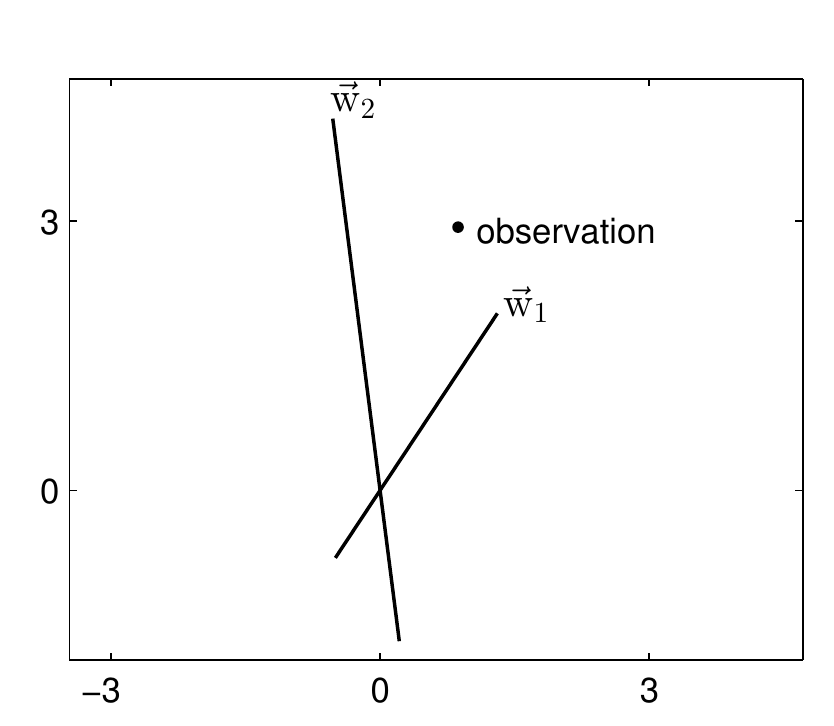}
\end{minipage}
\raisebox{2.05cm}{
\begin{minipage}{5cm}
\includegraphics[width=1.\textwidth]{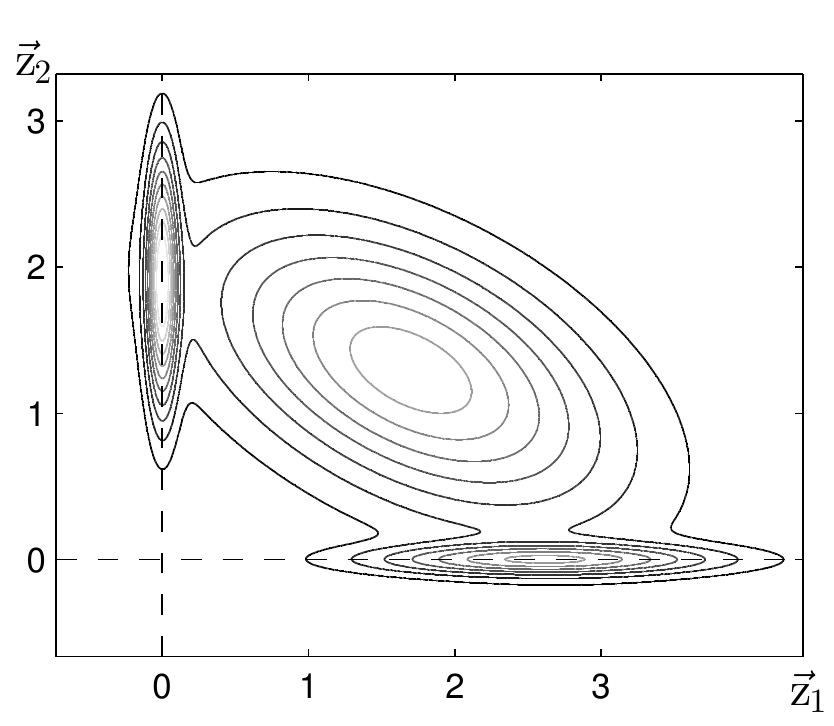}
\end{minipage} }
\caption{Left figures visualize observations generated by two different instantiations of the 
spike-and-slab sparse coding model \refp{EqnPriorS} to \refp{EqnNoise}. Solid lines are 
the generating bases vectors. Right figures illustrate the corresponding exact posteriors 
over latents computed using \refp{EqnPostS} and \refp{EqnAbbreviations} given observations 
and generating model parameters. The probability mass seen just along 
the axes or around the origin actually lies exactly on the axis.
Here we have spread the mass for visualization purposes by slightly augmenting zero diagonal entries of the posterior covariance matrix in 
\refp{EqnAbbreviations}.}
\label{fig:gsc_posteriors}
\end{center}
\end{figure}
The general challenge with spike-and-slab sparse coding models lies in the
optimization of the model parameters.
Whereas the standard Laplacian prior used for sparse coding results in uni-modal posterior distributions,
the spike-and-slab prior results in multi-modal posteriors~\citep[see, e.g.,][]{TitsiasLazaro2011,LuckeSheikh2012}.
Figure \ref{fig:gsc_posteriors} shows typical posterior distributions for spike-and-slab sparse coding 
(the model will be formally defined in the next section). The figure illustrates posterior examples for the case 
of a two-dimensional observed and a two-dimensional hidden space. As can be observed, 
the posteriors have multiple modes; and the number modes increases exponentially with the 
dimensionality of the hidden space \citep[][]{TitsiasLazaro2011,LuckeSheikh2012}. 
The multi-modal structure of the posteriors argues against the application of the 
standard maximum a-posteriori (MAP) approaches \citep{MairalBPS09,LeeEtAl2007,OlshausenField1997} or 
Gaussian approximations of the posterior \citep{Seeger2008,RibeiroOpper2011} because they rely on uni-modal
posteriors. 
The approaches that have been proposed in the literature are, consequently, MCMC based methods 
\citep[e.g.,][]{CarvalhoEtAl2008,ZhouetAl2009,MohamedEtAl2012} and variational EM methodologies 
\citep[e.g.,][]{ZhouetAl2009,TitsiasLazaro2011,GoodfellowEtAl2013}. While MCMC approaches are
more general and more accurate given sufficient computational
resources, variational approaches are usually more efficient.
Especially in high dimensional hidden spaces, the multi-modality of the posteriors is a particular challenge 
for MCMC approaches; consequently, recent applications to large hidden spaces have 
been based on variational EM optimization~\citep{TitsiasLazaro2011,GoodfellowEtAl2013}.
The variational approaches applied to spike-and-slab models thus far
\citep[see][]{RattrayEtAl2009,YoshidaEtAl2010,TitsiasLazaro2011,GoodfellowEtAl2013}
assume a factorization of the posteriors over the latent dimensions, that is 
the hidden dimensions are assumed to be independent a-posteriori.
This means that any dependencies such as explaining-away effects including correlations (compare Figure \ref{fig:gsc_posteriors}) 
are ignored and not accounted for. 
But what consequences does such a negligence
of posterior structure have? Does it result in biased parameter estimates and is it relevant for
practical tasks? Biases induced by factored variational inference in latent variable models have 
indeed been observed before
\citep{MacKay2001,IlinValpola2005,TurnerSahani2011}. For instance, in source separation tasks, 
optimization through factored inference can be biased towards finding mixing
matrices that represent orthogonal sparse directions, because such solutions are most
consistent with the assumed a-posteriori independence 
\citep[see][for a detailed discussion]{IlinValpola2005}. Therefore, the posterior independence assumption
in general may result in suboptimal solutions.

In this work we study an approximate EM approach for spike-and-slab
sparse coding which does not assume a-posteriori independence 
and which can model multiple modes. The novel approach can be considered as a variational EM approach but
instead of using factored distributions or Gaussians, it is based on posterior
distributions truncated to regions of high probability mass
\citep{LuckeEggert2010}. Such truncated EM approaches have recently been applied
to different models \citep[see e.g.,][]{PuertasEtAl2010,SheltonEtAl2011,DaiLucke2012a,BornscheinEtAl2013}.
In contrast to the previously studied factored variational approaches
\citep[][]{TitsiasLazaro2011,MohamedEtAl2012,GoodfellowEtAl2013}, the truncated
approach will furthermore take advantage of the fact that 
in the case of a Gaussian slab and Gaussian noise model, integrals
over the continuous latents can be obtained in closed-form
\citep{LuckeSheikh2012}. This implies that the posteriors over 
latent space can be computed exactly if the sums over the binary
part are exhaustively evaluated over exponentially many states. 
This enumeration of the binary part becomes
computationally intractable for high-dimensional hidden spaces.
However, by applying the truncated variational approach
exclusively to the binary part of the hidden space, we can still fully
benefit from the analytical tractability of the continuous integrals.

In this study, we systematically compare the truncated approach to a recently
suggested factored variational approach \citep{TitsiasLazaro2011}.  A
direct comparison of the two variational approaches will allow for
answering the questions about potential drawbacks and biases of both
optimization procedures. As approaches assuming factored variational
approximations have recently shown state-of-the-art performances
\citep{TitsiasLazaro2011,GoodfellowEtAl2013}, understanding their
strengths and weaknesses is crucial for further advancements of sparse
coding approaches and their many applications.  Comparison with other
approaches that are not necessarily based on the spike-and-slab model
will allow for accessing the potential advantages of the
spike-and-slab model itself.

In Section \ref{SecGSCmodel} we will introduce the used spike-and-slab
sparse coding generative model, and briefly discuss the factored variational
approach which has recently been applied for parameter optimization.
In Section \ref{SecEM} we derive the closed-form EM parameter update equations for the
introduced spike-and-slab model. Based on these equations, in Section \ref{sec:ET} we derive 
the truncated EM algorithm for efficient learning in high
dimensions. In Section \ref{sec:Experiments}, we numerically evaluate the algorithm 
and compare it to factored variational and other approaches. Finally, in Section \ref{Sec:Discussion} 
we discuss the results. The Appendix present details of the derivations and experiments.
\section{Spike-and-slab Sparse Coding} 
\label{SecGSCmodel}
The spike-and-slab sparse coding model assumes like standard sparse coding a
linear superposition of basis functions, independent latents, and 
Gaussian observation noise. The main difference is that a spike-and-slab
distribution is used as a prior. Spike-and-slab distributions have long been used 
for different models \citep[e.g.,][among many others]{MitchellBeauchamp1988}
and also variants of sparse coding with spike-and-slab priors have
been studied previously 
\citep[compare][]{West2003,GarriguesOlshausen2007,KnowlesGhahramani2007,TehGorurGhar2007,CarvalhoEtAl2008,PaisleyLawrence2009,ZhouetAl2009}.
In this work we study a generalization of the spike-and-slab sparse coding 
model studied by \citet{LuckeSheikh2012}.
The data generation process in the model assumes 
an independent Bernoulli prior for each component of the 
the binary latent vector  $\sVec \in \{0,1\}^H$ 
and a multivariate Gaussian prior for the continuous latent vector $\zVec \in \RRR^H$:
\begin{eqnarray}
p(\sVec\,|\Theta) &=& \Bernoulli(\sVec;\piVec) = \disS\prod_{h=1}^{H}\pi_h^{s_h}\,(1-\pi_h)^{1-s_h}, \label{EqnPriorS} \\
p(\zVec\,|\Theta) &=& \NGauss(\zVec;\,\muVec , \Sigmah)\label{EqnPriorZ},
\end{eqnarray}
where $\pi_h$ defines the probability of $s_h$ being equal to one and where $\muVec$ and $\Psi$
parameterize the mean and covariance of $\zVec$, respectively. The parameters 
$\muVec\in\RRR^H$ and $\Sigmah\in\RRR^{H \times H}$ parameterizing the Gaussian 
slab in \refp{EqnPriorZ} generalize the spike-and-slab model used in \citep{LuckeSheikh2012}.
A point-wise multiplication of the two latent vectors, i.e., $(\sVec\odot\zVec)_h = s_h\,z_h$ 
generates a `spike-and-slab' distributed variable ($\sVec\odot\zVec$), which 
has either continuous values or exact zero entries.
Given such a latent vector, a $D$-dimensional observation
$\yVec\in\RRR^{D}$ is generated by linearly 
superimposing a set of basis functions $W$ and 
by adding Gaussian noise:
\begin{align}
p(\yVec\,|\,\sVec,\zVec,\Theta) &= \NGauss(\yVec;\,W(\sVec\odot\zVec),\Sigmad),
\label{EqnNoise}
\end{align}
where each column of the matrix $W\in\RRR^{D \times H}$ is a basis function $W=(\vec{w}_1,\ldots,\vec{w}_H)$ 
and where the matrix $\Sigmad\in\RRR^{D \times D}$ parameterizes the observation noise.
Full rank covariances $\Sigmad$ can flexibly parametrize noise and have been found beneficial in 
noisy environments \citep{DalenGales2008,RanzatoHinton2010,DalenGales2011}. Nevertheless the model can also be constrained to have  
homoscedastic noise (i.e., $\Sigmad = \sig^2\One$).
We use $\Theta = (W,\Sigmad,\piVec,\muVec,\Sigmah)$ to denote all the model parameters. 
Having a spike-and-slab prior implies that for high levels of sparsity (low
values of $\pi_h$) the latents assume exact zeros with a high probability. 
This is an important distinction compared to the Laplace or Cauchy distributions
used for standard sparse coding \citep{OlshausenField1997}.

The spike-and-slab sparse coding algorithm we derive in this work is
based on the model \refp{EqnPriorS} to \refp{EqnNoise}. 
The factored variational approach 
\citep[Multi--Task and Multiple Kernel Learning, MTMKL;][]{TitsiasLazaro2011} 
that we use for detailed comparison is based on a similar model.  
The MTMKL model is both a constrained and generalized version of the model we study. 
On one hand, it is more constrained by assuming the same sparsity for each latent,
i.e., $\pi_h=\pi_{h^\prime}$ (for all $h,h^\prime$); 
and by using a diagonal covariance matrix for the observation noise,
$\Sigma=\mathrm{diag}(\sigma_1^2,\ldots,\sigma_D^2)$.  On the other
hand, it is a generalization by drawing the basis functions
$W$ from Gaussian processes. 
The model \refp{EqnPriorS} to
\refp{EqnNoise} can then be recovered as a special case of the MTMKL model if the Gaussian
processes are Dirac delta functions. 
For parameter optimization, the MTMKL model uses a 
standard factored variational optimization. In the case of spike-and-slab models, this factored approach means
that the exact posterior
$p(\sVec,\zVec\,|\,\yVec)$ is approximated by a variational
distribution $q_n(\sVec,\zVec;\,\Theta)$ which assumes the combined
latents to be independent a-posteriori
\citep[compare][]{ZhouetAl2009,TitsiasLazaro2011,GoodfellowEtAl2013}:
\begin{align}
\label{EqnFactoredPosterior}
q_n(\sVec,\zVec;\Theta)&= \disS\prod_{h=1}^{H}q_n^{(h)}(s_h,z_h;\Theta), \nonumber
\end{align}
where $q_n^{(h)}$ are distributions only depending on $s_h$ and $z_h$
and not on any of the other latents.
A detailed account of the MTMKL optimization algorithm is given by \citet{TitsiasLazaro2011}
and for later numerical experiments on the model, we used the source code provided along with that publication.\footnote{We downloaded the code from 
http://www.well.ox.ac.uk/\~mtitsias/code/varSparseCode.tar.gz.}
Further comparisons will include the spike-and-slab sparse coding model by \citet{ZhouetAl2009}.
The generative model is similar to the spike-and-slab model in Equations \refp{EqnPriorS} to \refp{EqnNoise}
but uses a Beta process prior to parameterize the Bernoulli (the ``spike'') distribution and assumes
homoscedastic observation noise.
Inference in their model is based on factored variational EM or Gibbs sampling. 
As this model is closely related to ours, we use it
as another instance for comparison in our numerical experiments in order to assess the influence of different inference method choices.
This comparison allows us to explore differences of training the model with a sampling-based approach, as they yield many of the same benefits of our inference method (e.g., flexible representation of uncertainty), but where generally more computational resources are necessary.

\section{Expectation Maximization for Parameter Optimization}
\label{SecEM}
In order to learn the model parameters $\Theta$ given a set of 
$N$ independent data points \mbox{$\{\yVecN\}_{n=1,\ldots,N}$} with $\yVecN\in\RRR^D$, 
we maximize the data likelihood
${\cal L}=\prod_{n=1}^N{}p(\yVecN\,|\,\Theta)$ by applying the Expectation
Maximization (EM) algorithm. Instead of directly maximizing the likelihood,
the EM algorithm \citep[in the form studied by][]{NealHinton1998} 
maximizes the free-energy, a lower bound of the log-likelihood given by:
\begin{equation}
\FF(\ThetaOld,\Theta) =
    \sum\limits_{n=1}^{N} \left\langle \log p(\yVecN, \sVec, \zVec\,|\,\Theta)
    \right\rangle_n + H(\ThetaOld),
\label{EqnFreeEnergy}
\end{equation}
where $\langle \,\cdot\, \rangle_n$ denotes the expectation under the posterior over the latents $\sVec$ and $\zVec$ given $\yVecN$
\begin{equation}
\E{f(\sVec,\zVec)}_n =
\sum_{\sVec}\int_{\zVec}\,p(\sVec,\zVec\,|\,\yVecN,\ThetaOld)\,
f(\sVec,\zVec)\,\dz
\label{EqnExpWRTPost}
\end{equation}
and  $H(\ThetaOld) = -\sum_{\sVec}\int_{\zVec}\,p(\sVec,\zVec\,|\,\yVecN,\ThetaOld)
\log(p(\sVec,\zVec\,|\,\yVecN,\ThetaOld))\,\dz$ 
is the Shannon entropy, which only depends on parameter
values held fixed during the optimization of $\FF$ w.r.t.\ $\Theta$ in the M-step. 
Here $\sum_{\sVec}$ is a summation over all possible binary vectors $\sVec$.

The EM algorithm iteratively optimizes the free-energy by alternating between two steps. First, in the E-step 
given the current parameters $\ThetaOld$, the relevant expectation values under the posterior 
$p(\sVec,\zVec\,|\,\yVecN,\ThetaOld)$ are computed. Next, the M-step uses these posterior expectations and maximizes 
the free-energy $\FF(\ThetaOld,\Theta)$ w.r.t.\ $\Theta$. Iteratively applying E- and M-steps
locally maximizes the data likelihood. In the following section we will first derive the
M-step equations which themselves will require expectation values over the posteriors \refp{EqnExpWRTPost}.
The required expressions and approximations for these expectations (the E-step) will be derived afterwards.
\subsection{M-step Parameter Updates}
The M-step parameter updates of the model are canonically obtained by
setting the derivatives of the free-energy \refp{EqnFreeEnergy}
w.r.t.\ the second argument to zero. Details of the derivations are
given in Appendix \ref{sec:UpdRulesDerv} and the resulting update equations
are as follows:
\begin{eqnarray}
\label{EqnMStepStart}
  W \ssb&=&\ssb \frac{\sum_{n=1}^{N} \yVecN \E{\sVec\odot\zVec}^{\TT}_n}
  {\sum_{n=1}^{N} \E{(\sVec\odot\zVec)(\sVec\odot\zVec)^{\TT}}_n }, \\
\piVec \ssb&=&\ssb
\frac{1}{N}\sum_{n=1}^{N}\E{\sVec}_n,\\
\muVec  \ssb&=&\ssb \frac{\sum_{n=1}^{N} \E{\sVec\odot\zVec}_n}{\sum_{n=1}^{N}\E{\sVec}_n}  \label{EqnMStepnu}, \\
\Sigmah \ssb&=&\ssb
    \sum_{n=1}^{N} \Big[\E{(\sVec\odot\zVec)(\sVec\odot\zVec)^{\TT}}_n 
    - \E{\sVec\,\sVec^{\TT}}_n \odot \muVec\muVec^{\TT} \Big] 
    \odot \Big(\sum_{n=1}^{N}\Big[\E{\sVec\,\sVec^{\TT}}_n \Big]\Big)^{-1},\\
\mbox{and} \ \     %
\Sigmad \ssb&=&\ssb \frac{1}{N}\sum_{n=1}^{N}\Big[ \yVecN(\yVecN)^{\TT}
        - W\big[\E{(\sVec\odot\zVec)}_n\E{(\sVec\odot\zVec)}_n^{\TT}\big]W^{\TT}\Big]\label{EqnMStepSigma}
    . \label{EqnMStepEnd}
\end{eqnarray}
\subsection{E-step Expectation Values}
\label{sec:Exact-E-step}
The M-step equations \refp{EqnMStepStart} to
\refp{EqnMStepEnd} require expectation values
w.r.t.\ the posterior distribution be computed over the whole latent space, which
requires either analytical solutions or approximations of integrals/sums
over the latent space.
For the derivation of closed-form E-step equations it is useful to know
that the discrete latent variable $\sVec$ can be combined with the
basis function matrix $W$ so that we can rewrite \refp{EqnNoise} as
\begin{align}
  p(\yVec\,|\,\sVec,\zVec,\Theta) &= \NGauss(\yVec;\,\Wt_{\sVec}\,\zVec,\Sigmad), \nonumber
%
\end{align}
where we have defined $(\Wt_{\sVec})_{dh}\,=\,W_{dh}s_h$ such that
\mbox{$W(\sVec\odot\zVec)\,=\,\Wt_{\sVec}\,\zVec$}.

Here the data likelihood $p(\yVec\,|\,\Theta)$ can be derived in closed-form
after marginalizing the joint $p(\yVec,\sVec,\zVec\,|\,\Theta)$ over $\zVec$:
\begin{align}
p(\yVec, \sVec |\,\Theta) &= \Bernoulli(\sVec;\piVec) \int
\NGauss(\yVec; \Wt_{\sVec}\,\zVec, \Sigmad) \, \NGauss(\zVec;\muVec, \Psi)\, \dz \nonumber \\
&= \Bernoulli(\sVec;\piVec)\ \NGauss(\yVec; \Wt_{\sVec}\,\muVec, C_{\sVec}), \label{EqnJointYS}
\end{align}
where $C_{\sVec}  = \Sigmad + \Wt_{\sVec}\,\Psi\Wt_{\sVec}^T$. 
The second step follows from standard identities for Gaussian random
variables \citep[e.g.,\,][]{Bishop2006}.
We can then sum the resulting expression over $\sVec$ to obtain
\begin{align}
p(\yVec\,|\,\Theta) &= \sum_{\sVec} \Bernoulli(\sVec;\piVec)\
\NGauss(\yVec; \Wt_{\sVec}\,\muVec, C_{\sVec}). \label{EqnMarginal}
\end{align}
Thus, the marginal distribution takes the form of a Gaussian mixture
model with $2^H$ mixture components indexed by $\sVec$.  However,
unlike in a standard Gaussian mixture model, the mixing
proportions and the parameters of the mixture components are not
independent but coupled together. Therefore, the following steps 
will lead to closed-form EM updates that are notably not a
consequence of closed-form EM for classical Gaussian mixtures. In contrast, 
Gaussian mixture models assume independent mixing proportions and independent component parameters.
By using Equations \refp{EqnJointYS} and \refp{EqnMarginal} the posterior
over the binary latents $p(\sVec\,|\,\yVec,\Theta)$ is given by:
\begin{equation}
  p(\sVec\,|\,\yVec,\Theta)  = \frac{p(\sVec,
  \yVec\,|\,\Theta)}{p(\yVec\,|\,\Theta)} = 
 \frac{\Bernoulli(\sVec;\piVec)\
 \NGauss(\yVec;\Wt_{\sVec}\,\muVec, C_{\sVec})}
 {\sum_{\sVec^{\prime}} \Bernoulli(\sVec^{\prime};\piVec)\
 \NGauss(\yVec;\,\Wt_{\sVec^{\prime}}\muVec, C_{\sVec^{\prime}})}. \label{EqnPostS}
\end{equation}
We can now consider the factorization of the posterior $p(\sVec,\zVec\,|\,\yVec,\Theta)$
into the posterior over the binary part $p(\sVec\,|\,\yVec,\Theta)$
and the posterior over the continuous part given the binary state
$p(\zVec\,|\,\sVec, \yVec, \Theta)$:
\begin{equation}
p(\sVec,\zVec\,|\,\yVec,\Theta) = p(\sVec\,|\,\yVec,\Theta)\,
p(\zVec\,|\,\sVec, \yVec, \Theta).\label{EqnPostSAndZ}
\end{equation}
Like the first factor in \refp{EqnPostSAndZ}, the second factor is
also analytically tractable and given by:
\begin{equation}
\begin{aligned}
  p(\zVec\, |\, \sVec, \yVec, \Theta) &=
  \frac{p(\sVec\,|\,\Theta)\,p(\zVec\,|\,\Theta)\,p(\yVec\, |\,
  \zVec, \sVec, \Theta)}{p(\sVec\,|\,\Theta) \int
  p(\yVec\,|\,\zVec, \sVec,\Theta)\,p(\zVec\,|\, \Theta)\dz}\nonumber \\
  &\propto \NGauss(\zVec; \muVec, \Psi) \,
  \NGauss(\yVec;\Wt_{\sVec}\,\zVec,\Sigmad)\\
  &= \NGauss(\zVec;\kappaVec_{\sVec}, \Lambda_{\sVec}),
\end{aligned}
\end{equation}
where the last step again follows from standard Gaussian identities with definitions
\begin{equation}
\begin{aligned}
\Lambda_{\sVec} &= (\Wt_{\sVec}^{\TT}\,\Sigmad^{-1}\,\Wt_{\sVec}\,+\,\Sigmah_{\sVec}^{-1})^{-1},\\ 
\kappaVecN_{\sVec} &= (\sVec\odot\muVec) + \Lambda_{\sVec}\,\Wt^{\TT}_{\sVec}\,\Sigmad^{-1}\,(\yVecN - \Wt_{\sVec}\,\muVec)
\label{EqnAbbreviations}
\end{aligned}
\end{equation}
and with $\Sigmah_{\sVec} = \Sigmah \big(\mathrm{diag}(\sVec)\big)$.
The full posterior distribution can thus be written as
\begin{eqnarray}
p(\sVec,\zVec\,|\,\yVecN,\Theta) 
&=&
\frac{\Bernoulli(\sVec;\piVec)\,\NGauss(\yVecN;\Wt_{\sVec}\,\muVec,C_{\sVec})\,\NGauss(\zVec;\,\kappaVecN_{\sVec},\Lambda_{\sVec})}
{\sum_{\sVec^{\prime}}\Bernoulli(\sVec^{\prime};\piVec)\,\NGauss(\yVecN;\Wt_{\sVec^{\prime}}\,\muVec,C_{\sVec^{\prime}})}.
\label{EqnPostMain} 
\end{eqnarray}
Equation \refp{EqnPostMain} represents the crucial
result for the computation of the E-step below because, first, it
shows that the posterior does not involve analytically intractable
integrals and, second, for fixed $\sVec$ and $\yVecN$ the dependency
on $\zVec$ follows a Gaussian distribution. This special form allows
for the derivation of analytical expressions for the expectation
values as required for the M-step updates. Because of the Gaussian
form, the integrations over the continuous part are straight-forward
and the expectation values required for the M-step are given as follows:
\begin{eqnarray}
\E{\sVec}_n &=& \sum_{\sVec}q_n(\sVec;\Theta)\,\sVec{},\label{EqnEStepStart}\\
\E{\sVec\,\sVec^{\TT}}_n &=& \sum_{\sVec}q_n(\sVec;\Theta)\,\sVec\,\sVec^{\TT},\\ 
\E{\sVec\odot\zVec}_n &=& \sum_{\sVec}q_n(\sVec;\Theta)\,\kappaVecN_{\sVec}\label{EqnEStep},\\
\mbox{\ \ and\ \ }
\E{(\sVec\odot\zVec)(\sVec\odot\zVec)^{\TT}}_n 
&=&
\sum_{\sVec}q_n(\sVec;\Theta)\,\big(\Lambda_{\sVec} + \kappaVecN_{\sVec}(\kappaVecN_{\sVec})^{\TT}\big).\label{EqnEStepEnd}
\end{eqnarray}
In all of the expressions above, the left-hand-sides are expectation values
over the full latent space w.r.t.\ the posterior $p(\sVec,\zVec\,|\,\yVecN,\Theta)$,
whereas the right-hand-sides now take the form of expectation values only over
the binary part w.r.t.\ the posterior $p(\sVec\,|\,\yVecN,\Theta)$ in Equation \refp{EqnPostS}.
The derivations of E-step equations \refp{EqnEStepStart} to
\refp{EqnEStepEnd} are a generalization of the derivations by
\citet{LuckeSheikh2012}. While Gaussian identities and
marginalization have been used to obtain analytical results for
mixture-of-Gaussians priors before
\citep[e.g.][]{MoulinesEtAl1997,Attias1999,OlshausenMillman2000,GarriguesOlshausen2007}, the
above equations are the first closed-form solutions for the
spike-and-slab model \citep[first appearing in][]{LuckeSheikh2012}.
The observation that the Gaussian slab and Gaussian noise model allows for
analytically tractable integrals has, in parallel work, also been
pointed out by \citet[][]{MohamedEtAl2012}.

Iteratively computing the E-step equations \refp{EqnEStepStart} to
\refp{EqnEStepEnd} using the current
parameters $\Theta$ and the M-step equations \refp{EqnMStepStart} to
\refp{EqnMStepEnd}, represents a
closed-form and exact EM algorithm which increases the data likelihood
of the model to (possibly local) maxima. 
\section{Truncated EM}
\label{sec:ET}

While being exact, the execution of the above EM algorithm results in
considerable computational costs for larger-scale problems.  Without
approximations, the computational resources required scale
exponentially with the number of hidden dimensions $H$.  This can be
seen by considering the expected values w.r.t.\ the posterior
$p(\sVec\,|\,\yVec,\Theta)$ above, which each require a summation over
all binary vectors $\sVec\in \{0,1\}^H$. For tasks involving low
dimensional hidden spaces, the exact algorithm is still applicable.
For higher dimensional problems approximations are required, however.
Still, we can make use of the closed-form EM solutions by applying an
approximation solely to the binary part.  Instead of sampling-based or
factored approximations to the posterior
$p(\sVec,\zVec\,|\,\yVec,\Theta)$, we use a truncated
approximation to the posterior $p(\sVec\,|\,\yVecN,\Theta)$ in
Equation \refp{EqnPostS}. The truncated approximation is defined to
be proportional to the true posteriors on subspaces of the latent
space with high probability mass
\citep[compare {\em Expectation Truncation},][]{LuckeEggert2010}. More concretely, a posterior
distribution $p(\sVec\,|\,\yVecN,\Theta)$ is approximated by a
distribution $q_n(\sVec;\Theta)$ that only has support on a subset
$\KKn \subseteq \{0,1\}^H$ of the state space:
\begin{eqnarray}
\label{EqnETPosterior}
q_n(\sVec;\Theta) = \frac{p(\sVec,\yVecN\,|\,\Theta)}{\disS\sum_{\sVecPrime\in\KKn}p(\sVecPrime,\yVecN\,|\,\Theta)}\,\delta(\sVec\in\KKn),
\end{eqnarray}
where $\delta(\sVec\in\KKn)$ is an indicator function, i.e., $\delta(\sVec\in\KKn)=1$ if $\sVec\in\KKn$ and zero otherwise.

\begin{figure}[t]
\begin{center}
\def\svgwidth{\textwidth}
\includegraphics[width=\textwidth]{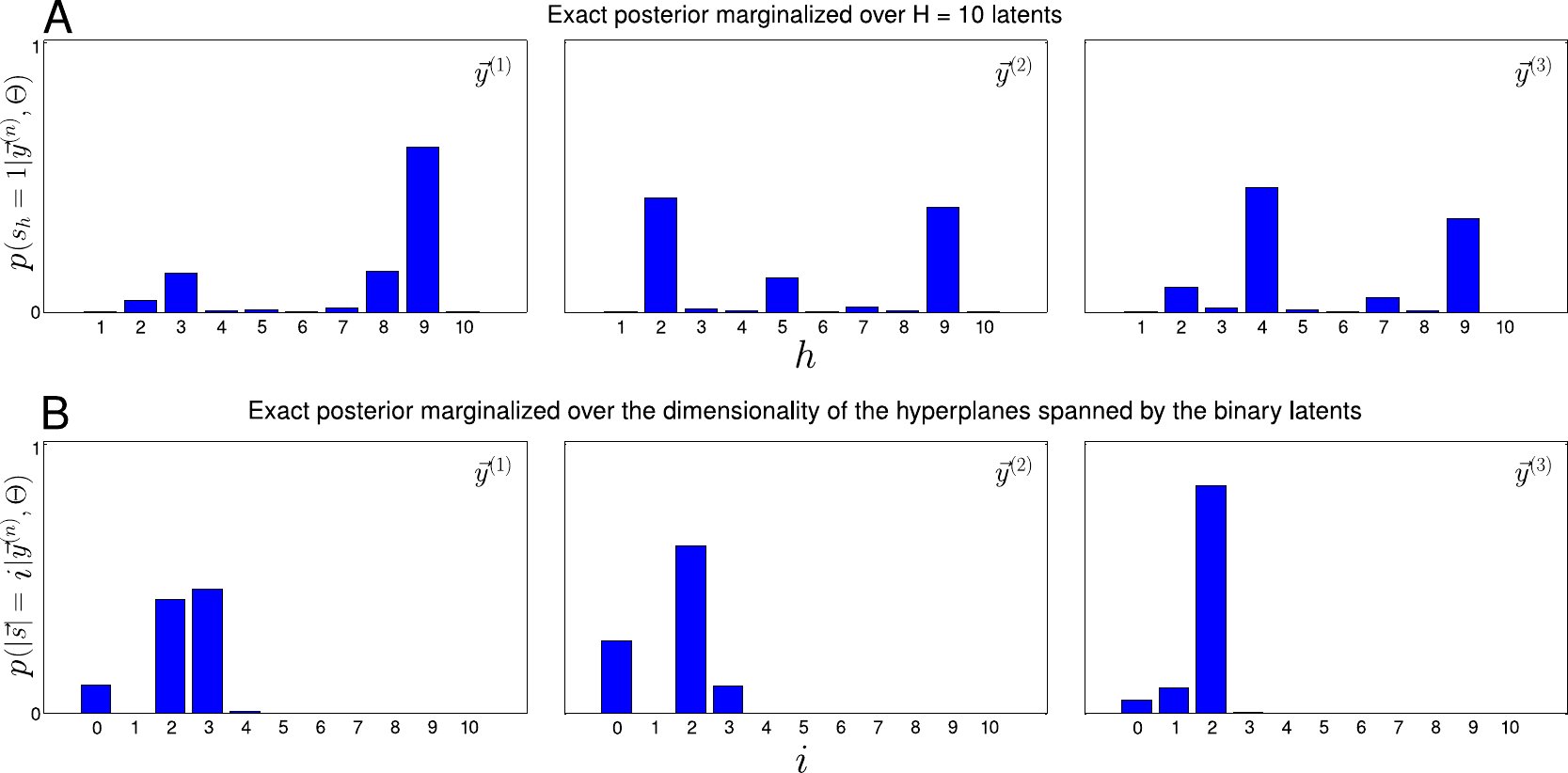}
\caption{
Visualization of the exact posterior probabilities of the spike-and-slab model with 
$H=10$ latents, computed for three given data points $\yVecN$.  
The model was trained on natural data (see Section \ref{sec:source-sep} for more details). 
{\bf A} Histograms of the posterior mass over the $H$ 
latents: $p(s_h=1\,|\,\vec{y}^{(n)},\Theta)=\sum_{\vec{s}_{\mathrm{with}}s_h=1}p(\vec{s}\,|\,\vec{y}^{(n)},\Theta)
/\sum_{\vec{s}}p(\vec{s}\,|\,\vec{y}^{(n)},\Theta)$.
Low values for most $h$ imply that these latents can be neglected (i.e., clamped to zero) for a posterior approximation.
{\bf B} Histograms of the posterior mass over the hyperplanes of increasing dimensionality $i$:
$p(|\sVec|=i\,|\,\vec{y}^{(n)},\Theta)=\sum_{\vec{s}_{\mathrm{where}}|\vec{s}|=i}p(\vec{s}\,|\,\vec{y}^{(n)},\Theta)
/\sum_{i^\prime= 0}^H\sum_{\vec{s}_{\mathrm{where}}|\vec{s}|=i^\prime}p(\vec{s}\,|\,\vec{y}^{(n)},\Theta)$. 
In case of all the three examples presented here, 
subspaces with $i > 4$ can be neglected as another approximation step for posterior estimation.
}\label{fig:sparse-post}
\end{center}
\end{figure}
The basic assumption behind the approximation in \refp{EqnETPosterior} is that the posterior
over the entire hidden space is concentrated in small volumes, which is represented by the reduced support of subset $\KKn$. 
When using a spike-and-slab sparse coding model to gain a generative understanding of the data, sparsity in the posterior distribution usually emerges naturally. We can see an illustration of this in 
Figure \ref{fig:sparse-post} (generation details in Section~\ref{sec:source-sep}).
Figure \ref{fig:sparse-post}A shows (for three typical data points) how much posterior mass is carried by each of the 
$H=10$ latent dimensions. Figure \ref{fig:sparse-post}B shows (for the same data points) histograms of the posterior 
mass marginalized across the whole range of hyperplanes spanned by the $10$--dimensional latent space.  
Figure \ref{fig:sparse-post}A indicates that only 
a subset of the $H$ latents is significantly relevant for encoding the posterior, while Figure \ref{fig:sparse-post}B 
allows us to observe that the posterior mass is primarily contained within low-dimensional hyperplanes of the $H$--dimensional hidden space. 
In other words, given a data
point we find that most of the posterior mass is concentrated in low-dimensional subspaces spanned by $H'\ll{}H$ of the latent dimensions. 
The sparse nature of the posterior as illustrated by Figure \ref{fig:sparse-post} allows us to apply approximation \refp{EqnETPosterior}, 
where we define the subsets $\KKn$ based on index sets
$I_n\subseteq\{1,\ldots,H\}$, which contain the indices of $H'$ most relevant sparse latents (compare Figure \ref{fig:sparse-post}A) 
for the corresponding data points $\yVecN$:
\begin{equation}
\label{EqnKKn}
\disT\KKn\, =\, 
\{\sVec\ |\ \sum_h{}s_h\leq{}\gamma\ \mbox{and}\ \forall{}h \not\in{}I_n: s_h=0 \}\,\cup\,{\cal U},
\end{equation}
where the indices comprising $I_n$ have the highest value 
of a selection (or scoring) function $\SCal_h(\yVecN,\Theta)$ (which we define
later). The set ${\cal U}$ is defined as ${\cal U}=\{\sVec\ |\ \sum_h{}s_h=1\}$ and ensures
that $\KKn$ contains all singleton states \citep[compare][]{LuckeEggert2010}. Otherwise,
$\KKn$ only contains vectors with at most $\gamma$ non-zero entries and with non-zero entries
only permitted for $h\in{}I_n$. The parameter $\gamma \leqslant H^{\prime}$ sets the maximal dimensionality of the considered hyper-planes
(compare Figure \ref{fig:sparse-post}B).
It was empirically shown by \citet{LuckeEggert2010} that for appropriately defined
subspaces $\KKn$, the KL-divergence between the true posteriors and
their truncated approximations converges to values close to zero.  

If we now use the concrete expressions of the sparse spike-and-slab model
(Equations \refp{EqnPriorS} to \refp{EqnNoise}) for the variational distribution
in Equation \refp{EqnETPosterior}, the truncated approximation is given by: 
\begin{eqnarray}
p(\sVec\,|\,\yVecN,\Theta)\ \approx\ q_n(\sVec;\Theta) 
&=& \frac{\NGauss(\yVecN;\Wt_{\sVec}\,\muVec,C_{\sVec})\,\Bernoulli(\sVec;\piVec)}
{\sum_{\sVecPrime\in\KKn}\NGauss(\yVecN;\Wt_{\sVec^{\prime}}\muVec,C_{\sVec^{\prime}})\,\Bernoulli(\sVec^{\prime};\piVec)\,}\,\delta(\sVec\in\KKn).
\label{eq:ET-GSC-post}
\end{eqnarray}
The approximation can now be used to
compute the expectation values which are required for the M-step
equations.  If we use the variational distributions in Equation \refp{eq:ET-GSC-post} for $q_n(\sVec;\Theta)$ on the
right-hand-sides of Equations \refp{EqnEStepStart} to \refp{EqnEStepEnd}, we
obtain:
\begin{eqnarray}
\sum_{\sVec}q_n(\sVec;\Theta)\,f(\sVec) &=& \frac{\sum_{\sVec\in\KKn}\NGauss(\yVecN;\Wt_{\sVec}\,\muVec,C_{\sVec})\,\Bernoulli(\sVec;\piVec)\,f(\sVec)}
{\sum_{\sVecPrime\in\KKn}\NGauss(\yVecN;\Wt_{\sVec^{\prime}}\muVec,C_{\sVec^{\prime}})\,\Bernoulli(\sVec^{\prime};\piVec)\,},
\label{eq:ET-GSC-expect}
\end{eqnarray}
where $f(\sVec)$ denotes any of the (possibly parameter dependent) functions of \refp{EqnEStepStart} to \refp{EqnEStepEnd}.
Instead of having to compute sums over the entire binary state space with $2^H$ states,
only sums over subsets $\KKn$ have to be computed. 
Since for many applications the posterior mass is finally
concentrated in small volumes of the state space, the approximation
quality can stay high even for relatively small sets $\KKn$.

Note that the definition of $q_n(\sVec;\Theta)$ in
Equation \refp{EqnETPosterior} neither assumes uni-modality like MAP approximations
\citep{MairalBPS09,LeeEtAl2007,OlshausenField1997} or 
Gaussian approximations of the posterior \citep{RibeiroOpper2011,Seeger2008}, 
nor does it assume a-posteriori independence of
the latents as factored approximations
\citep{JordanEtAl1999,GoodfellowEtAl2013,TitsiasLazaro2011}. 
The approximation scheme we have introduced here 
exploits the inherent property of the sparse spike-and-slab model
to have posterior probabilities concentrated in low-dimensional subspaces.
The quality of our approximated posterior $q_n(\sVec;\Theta)$ primarily depends on an appropriate
selection of the relevant subspaces $\KKn$ (see Section \ref{sec:SelectionFunction} below).

The truncated approximation is similar to factored variational approximations or MAP approximations
in the sense that it can be formulated as an approximate distribution $q_n(\sVec;\Theta)$ within the
free-energy formulation by \cite{NealHinton1998}. Within this formulation, $q_n(\sVec;\Theta)$ is
often referred to as {\em variational} approximation, and we therefore refer to our approximation
as {\em truncated variational EM}. Like factored variational approaches, we here aim to minimize
the KL-divergence between the true posterior and the approximation in Equation \refp{EqnETPosterior}.
However, we do not use variational parameters and a gradient based optimization of such parameters
for the minimization. Our approach is therefore not a variational approach in the sense of classical
variational calculus.
\subsection{Computational Complexity}
\label{sec:ET-complexity}
The truncated E-step defined by
\refp{EqnEStepStart} to \refp{EqnEStepEnd} with \refp{eq:ET-GSC-expect} scales 
with the approximation parameters $\gamma$ and $H^{\prime}$ which can be defined
independently of the latent dimensionality $H$. The complexity scales as
$\BigO \big(N\sum_{\gamma^{\prime}=0}^{\gamma}{H^{\prime}\choose \gamma^{\prime}}(D + \gamma^{\prime})^3\big)$, where the $D^3$ term can be
dropped from the cubic expansion if the observed noise $\Sigmad$ is considered to be diagonal or homoscedastic.
Also the truncated approximation yields sparse matrices in 
Equations \refp{EqnETPosterior} and \refp{eq:ET-GSC-post}
which results in more efficient and tractable matrix operations.

Although the total number of data points $N$ above defines a theoretical upper bound, 
in practice we can further benefit from the preselection step of the truncated approach to 
achieve significantly improved runtime performances. Clustering the data points using the index sets $I_n$ 
saves us from redundantly performing various computationally expensive operations involved 
in Equations \refp{EqnAbbreviations} and \refp{eq:ET-GSC-post}, that given a state $\sVec\in\KKn$ are 
independent of individual data points sharing the same subspace $\KKn$. 
Furthermore, such a batch processing strategy is also readily 
parallelizable as the truncated E-step can be performed independently for individual data clusters
(see Appendix \ref{sec:ParallelProcessing} for details).
Using the batch execution mode we have observed an average 
runtime speedup of up to an order of magnitude.
\subsection{Selection Function}
\label{sec:SelectionFunction}
To compose appropriate subspaces $\KKn$ a selection 
function $\SCal_h(\yVecN,\Theta)$ is defined, which prior to each 
E-step allows us to select the relevant $H^{\prime}$ hidden 
dimensions (i.e., the elements of the index
sets $I_n$) for a given observation $\yVecN$. A selection function is essentially a 
heuristic-based scoring 
mechanism, that ranks all the latents based on their potential for 
being among the generating causes of a given observation. Selection functions can
be based on upper bounds for probabilities $p(s_h=1\,|\,\yVecN,\Theta)$ 
\citep[compare][]{LuckeEggert2010,PuertasEtAl2010} or 
deterministic functions such as the scalar product between a basis vector and 
a data point \citep[derived from noiseless 
limits applied to observed space; compare][]{LuckeEggert2010,BornscheinEtAl2013}.

For the sparse coding model under consideration we define a selection function as follows:
\begin{eqnarray}
\Scal_h(\yVecN,\Theta) \ssb&=&\ssb \NGauss(\yVecN;\,\Wt_{\sVec_h}\muVec, C_{\sVec_h}) 
		\propto p(\yVecN\,|\,\sVec = \sVec_h,\Theta),
\label{eq:sel-func}
\end{eqnarray}
where $\sVec_h$ represents a singleton state in which only the entry $h$ is non-zero. 
The selection function \refp{eq:sel-func} is basically the data likelihood 
given a singleton state $\sVec_h$. The function does not take into account
the probability of the state itself (i.e., $p(\sVec_h\,|\,\Theta)$), as this may 
introduce a bias against less active latent dimensions. 
Similar to previously used selection
functions \citep[compare e.g.,][]{LuckeEggert2010,PuertasEtAl2010}, 
in order to maintain a linear scaling behavior w.r.t.\ the number of latents, 
the selection function introduced here avoids computationally demanding 
higher-order combinatorics of the latents by only assessing one-to-one correspondences between individual latents and
an observed data point. In the next section we empirically evaluate the efficacy of our selection function 
by means of numerical experiments that are based on the KL-divergence between the exact 
and the approximated posteriors computed from the subspaces $\KKn$.

Equations \refp{EqnETPosterior} to \refp{eq:ET-GSC-post} replace the
computation of the expectation values w.r.t.\ the exact posterior, and
represent the approximate EM algorithm used in the experiments section. The algorithm will
be applied without any further mechanisms such as annealing as we
found it to be very robust in the form derived above. Furthermore,
no data preprocessing such as mean subtraction or variance normalization
will be used in any of the experiments. To distinguish
the algorithm from others in comparative experiments, we will refer to it
as {\em Gaussian Sparse Coding} (GSC) algorithm in order to emphasize the
special Gaussian case of the spike-and-slab model used.
\section{Numerical Experiments}
\label{sec:Experiments}
We investigate the performance of the GSC algorithm on artificial data as well as various realistic 
source separation and denoising benchmarks. For all experiments the algorithm was implemented to
run in parallel on multiple CPUs with no dependency on their 
arrangement as physically collocated arrays with shared memory or distributed among multiple compute 
nodes \citep[see][for more details]{BornscheinEtAl2010}. 
We further extended the basic technique to make our implementation more efficient and suitable for 
parallelization by applying the batch execution 
(the observation discussed in Section \ref{sec:ET-complexity} on Computational Complexity and Appendix \ref{sec:ParallelProcessing}).  
In all the experiments, the GSC model parameters were randomly initialized.\footnote{We randomly and uniformly 
initialized the $\pi_h$ between $0.05$ and $0.95$. $\muVec$ was initialized with normally distributed random values and 
the diagonal of $\Sigmah$ was initialized with strictly positive uniformly distributed random values. 
We set $\Sigmad$ to the covariance across the data points, 
and the elements of $W$ were independently drawn from a normal distribution
with zero mean and unit variance.} The choice of GSC truncation parameters 
$H^{\prime}$ and $\gamma$ is in general straight-forward: the larger they are
the closer the match to exact EM but the higher are also the computational costs.
The truncation parameters are therefore capped by the available computational resources. 
However, empirically we observed that often much smaller values were sufficient than those that are maximally affordable.\footnote{Compare 
Appendix \ref{sec:PerfvsComp} for trade-off between complexity and accuracy of the truncated EM approach.}
Note that factored variational approaches do not usually offer such a trade-off between the 
exactness and computational demand of their inference schemes by means of a simple parameter adjustment.
\subsection{Reliability of the Selection Function}
To assess the reliability of the selection function we perform 
a number of experiments on small scale artificial data generated by the model, such that 
we can compute both the exact \refp{EqnPostS} and truncated \refp{eq:ET-GSC-post} 
posteriors. To control for the quality of the truncated posterior approximation---and thus the 
selection function---we compute the ratio between posterior mass within the truncated space $\KKn$
and the overall posterior mass \citep[compare][]{LuckeEggert2010}:
\begin{equation}
\Qn\,=\,
\frac{\sum_{\sVec\in\KKn}\int_{\zVec}p(\sVec,\zVec\,|\,\yVecN,\Theta)\,\dz}
{\sum_{\sVec^{\,\prime}}\int_{\zVec^{\prime}}p(\sVec^{\,\prime},\zVec^{\prime}\,|\,\yVecN,\Theta)\,\dz^{\prime}}\,
\,=\,\frac{\sum_{\sVec\in\KKn}\Bernoulli(\sVec;\piVec)\,\NGauss(\yVecN;\Wt_{\sVec}\,\muVec,C_{\sVec})\,}
{\sum_{\sVec^{\prime}}\Bernoulli(\sVec^{\prime};\piVec)\,\NGauss(\yVecN;\Wt_{\sVec^{\prime}}\muVec,C_{\sVec^{\prime}})},
\label{EqnQValues}
\end{equation}
where the integrals over the latent $\zVec$ in \refp{EqnQValues} are again given in closed-form.
The metric $\Qn$ ranges from zero to one and is directly related to the KL-divergence
between the approximation $q_n$ in Equation \refp{eq:ET-GSC-post} and the true posterior:
\begin{equation}
\DKL(q_n(\sVec,\zVec;\,\Theta),p(\sVec,\zVec\,|\,\yVecN,\Theta))\,=\,-\log(\Qn)\,. \nonumber
\end{equation}
If $\Qn$ is close to one, the KL-divergence is close to zero.

Data for the control experiments were generated by linearly superimposing  
basis functions that take the form of horizontal and vertical
bars \citep[see e.g.,][]{Foeldiak1990,Hoyer2002} on a $D = D_2\times D_2$ pixel grid, where $D_2 = H/2$. This gives us $D_2$ 
possible horizontal as well as vertical locations for bars of length $D_2$,
which together form our generating bases $W^{\mathrm{gen}}$. 
Each bar is then randomly assigned either a positive or negative value with magnitude $10$.
We set the sparsity such that there are on average two active bars per data point,
i.e., $\pi_h^{\mathrm{gen}}=2/H$ for all $h \in H$. We assume homoscedastic\footnote{To infer homoscedastic 
noise we set in the M-step the updated noise matrix $\Sigmad$ to $\sigma^2\One_D$ where $\sigma^2 = \trace\big(\Sigmad\big)/D$. 
This is equivalent to parameter update for $\sigma^2$ if the model originally assumes homoscedastic noise.}
observed noise $\Sigmad^{\mathrm{gen}} = \sig^2{\One_D}$, where $\sig^2=2.0$. 
The mean of the generating slab is i.i.d.\ drawn from a Gaussian: $\muVec^{\mathrm{gen}}\sim\NGauss(0,5)$, 
and the covariance of the slab is $\Sigmah^{\mathrm{gen}} = {\One_H}$.
We generate $N=1000$ data points. 
We run experiments with different sets of values for the truncation 
parameters $(H^{\prime},\gamma) \in \{(4,4),(5,4),(5,3)\}$ 
for each $H \in \{10,12\}$. Each run consists of $50$ EM iterations and after each run we compute
the Q-value over all the data points. For all the experiments we find the average Q-values to be 
above $0.99$, which shows that the state subspaces \refp{EqnKKn} constructed from the $H^{\prime}$ 
latents chosen through the selection function \refp{eq:sel-func} contain almost the entire 
posterior probability mass in this case. The small fraction of remaining posterior mass
lies in other discrete subspaces and its principle form is known to not contain any heavy tails
(see Equation \refp{EqnPostMain}). The contribution of the truncated posterior mass to parameter updates can
therefore be considered negligible.

\subsection{Consistency}

Prior to delving into a comparative analysis of GSC with other methods, 
we assess the consistency of the approach by applying both its 
exact and truncated variational inference schemes on the task of recovering 
sparse latent directions w.r.t.\ increasing numbers of training data. 
For this experiment we work with synthetic data generated by 
the GSC model itself. Moreover, we also apply the truncated variational inference  
on standard sparse coding data generated with a standard Laplace prior\citep{OlshausenField1996,LeeEtAl2007}. 
Taking into account the computational demands of the exact inference, we set 
both the hidden as well as observed dimensions ($H$ and $D$ respectively) to $10$. For the 
experiment we exponentially increase $N$ from $1000$ to $512000$. 
For each trial in the experiment we generate a new ground-truth  
mixing matrix $W^{\mathrm{gen}} \in \RRR^{D \times H}$ by randomly generating a set of $H$ orthogonal bases 
and perturbing them with a Gaussian noise with zero mean and a variance of $2.0$. 
We set the sparsity parameters $\pi_h$ to $1/H$, while the observed noise is assumed to be 
homoscedastic with $\sig = 1.0$. When generating data with a spike-and-slab prior,  
the slab is considered to have its mean at zero with an identity   
covariance matrix, i.e., $\mu_h = 0.0$ for all $h \in H$ and $\Sigmah^{\mathrm{gen}} = {\One_H}$, respectively.
In each trial after performing $100$ EM iterations and inferring the whole set of GSC parameters $\Theta$, 
we quantify the quality of the inference in terms of how well the inferred bases $W$ align 
with the corresponding ground truth bases $W^{\mathrm{gen}}$.
As a  measure of discrepancy between the generating and the recovered bases we use the 
Amari index \citep{AmariEtAl1996}:
\begin{eqnarray}
\disS{}A(W)&=&\disS \frac{1}{2H(H-1)} \sum_{h,h^\prime=1}^{H} \Big(  \frac {\vert O_{hh^\prime}\vert} {\max_{h^{\prime\prime}}\vert O_{hh^{\prime\prime}}\vert}
  +\,  \frac{ \vert O_{ hh^{\prime}}\vert } { \max_{h^{\prime\prime}}\vert O_{h^{\prime\prime}h^{\prime}}\vert } \Big)
- \frac{1}{H-1}, 
\label{EqnAmari}
\end{eqnarray}
where $O_{hh'} = \left(W^{-1}W^{\mathrm{gen}}\right)_{hh'}$. 
The Amari index is either positive or zero. 
It is zero only when the basis vectors of $W$ and $W^{\mathrm{gen}}$ 
represent the same set of orientations, which in our case implies a precise  
recovery of the (ground truth) sparse directions. 

\begin{figure}
\begin{center}
\includegraphics[width=.65\textwidth]{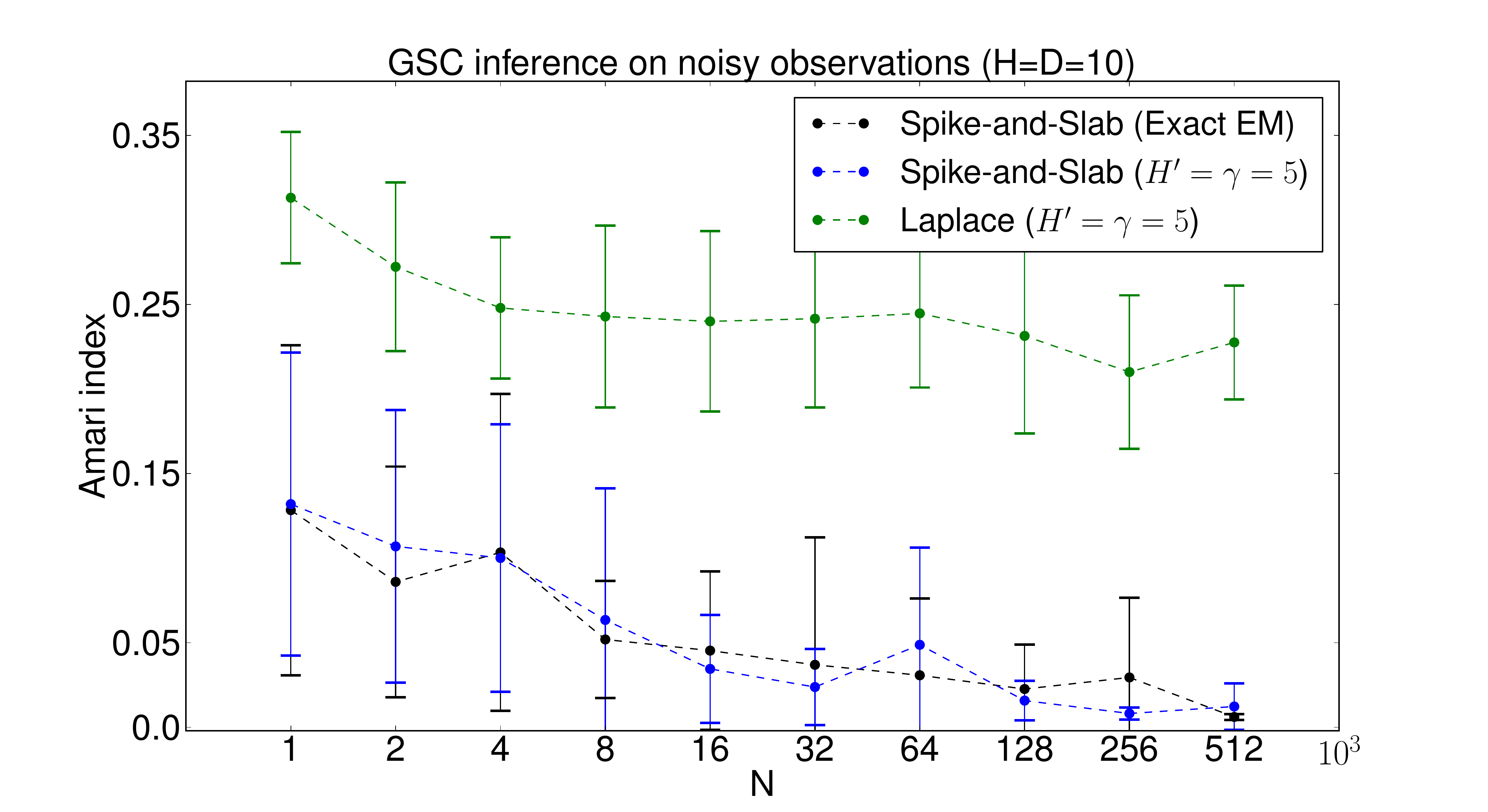}
\caption{Numerical experiment investigating the consistency of the exact as well as 
the truncated variational GSC algorithm for increasing
numbers of data points. The curves show results for the recovery of sparse
directions for different numbers of data points. Data points were generated by both the
spike-and-slab generative model (black and blue) and a standard sparse coding
model with Laplace prior (green). The curves show the mean
Amari index and standard deviations computed based on 15 repetitions of the learning
algorithm.
}
\label{fig:GSC-consistency}
\end{center}
\end{figure}

Figure \ref{fig:GSC-consistency} summarizes the results of the experiment. Each error bar in 
the plot extends one standard deviation on both sides of its corresponding mean Amari index, which 
is computed from $15$ repetitions. The black curve shows the results of the exact GSC inference on 
spike-and-slab generated data, while the blue and green curves illustrate the results of the 
truncated variational inference ($H^{\prime}=\gamma=5$) on data generated by spike-and-slab and 
Laplace priors respectively. For data generated with the spike-and-slab prior, we   
observe a gradually more accurate recovery of the sparse directions, as the mean 
Amari indices gradually converge towards the minimum value of zero for increasing numbers of training data. 
The minimum Amari index values that we obtain for the black and blue curves for $N \in \{128K, 256K, 512K\}$ 
are all below $6 \times 10^{-3}$. For the standard sparse coding data, we also see an improvement in performance with more data; 
however, higher mean values of the Amari index in this case can presumably be attributed to the model mismatch.

\subsection{Recovery of Sparse Directions on Synthetic Data}
In our first comparison with other methods, we measure the performances of GSC (using the truncated variational approximation)
and MTMKL (which uses a factored variational approximation) approaches 
on the sparse latent direction recovery task 
given synthetic data generated by standard sparse coding models. In one set
of experiments we generate data using sparse coding with Cauchy prior \citep{OlshausenField1996},
and in another set of experiments we use the standard Laplace distribution as a prior
\citep{OlshausenField1996,LeeEtAl2007}.
For each trial in the experiments a new 
mixing matrix $W^{\mathrm{gen}}$ was generated without any constraints 
on the sparse directions (i.e., matrices were non-orthogonal in general).
In both sets of experiments we simultaneously vary both the observed and latent dimensions $D$ 
and $H$ between $20$ and $100$, and repeat $15$ trials per given dimensionality. 
For each trial we randomly generated a new data set of $N=5000$ noisy 
observations with $\Sigmad^{\mathrm{gen}} = \One_D$. Per trial, we perform $50$ iterations of both 
algorithms. The GSC truncation parameters $H^{\prime}$ and $\gamma$ were set to $\frac{H}{10}$.
We assess the performances of the algorithms w.r.t.\ the Amari index \refp{EqnAmari}.

\begin{figure}
\begin{center}
\includegraphics[width=.85\textwidth]{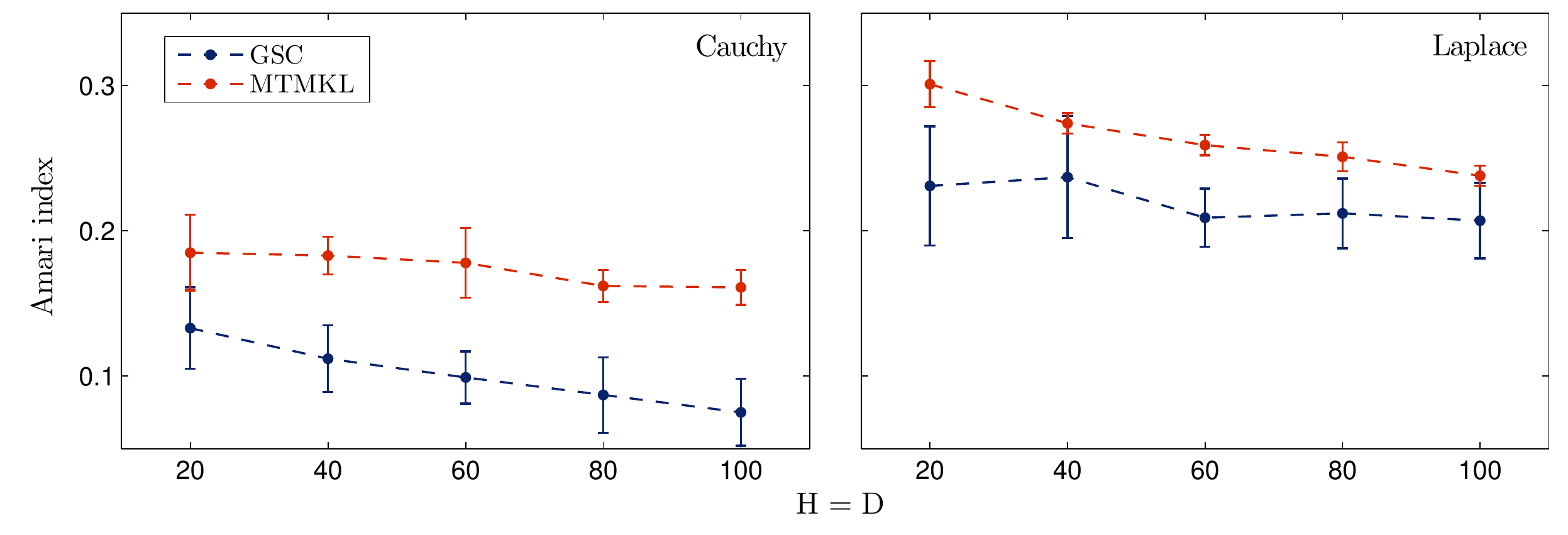}
\caption{Performance of GSC (with $H^{\prime} = \gamma = \frac{H}{10}$) vs.\ MTMKL
on data generated by standard sparse coding models both with Cauchy and Laplace priors. 
Performance compared on the Amari index \refp{EqnAmari}.
}
\label{fig:GSC-ET_vs_MTMKL_non-ortho}
\end{center}
\end{figure}

The results for GSC and MTMKL in Figure \ref{fig:GSC-ET_vs_MTMKL_non-ortho} show that both approaches
do relatively well in recovering the sparse directions, which shows that they are robust
against the model mismatch imposed by generating from models with other priors.
Furthermore, we observe that the GSC approach consistently recovers the sparse directions more accurately.

\subsection{Source Separation} 
\label{sec:source-sep}
\begin{figure}
\begin{center}
\includegraphics[width=0.99\textwidth]{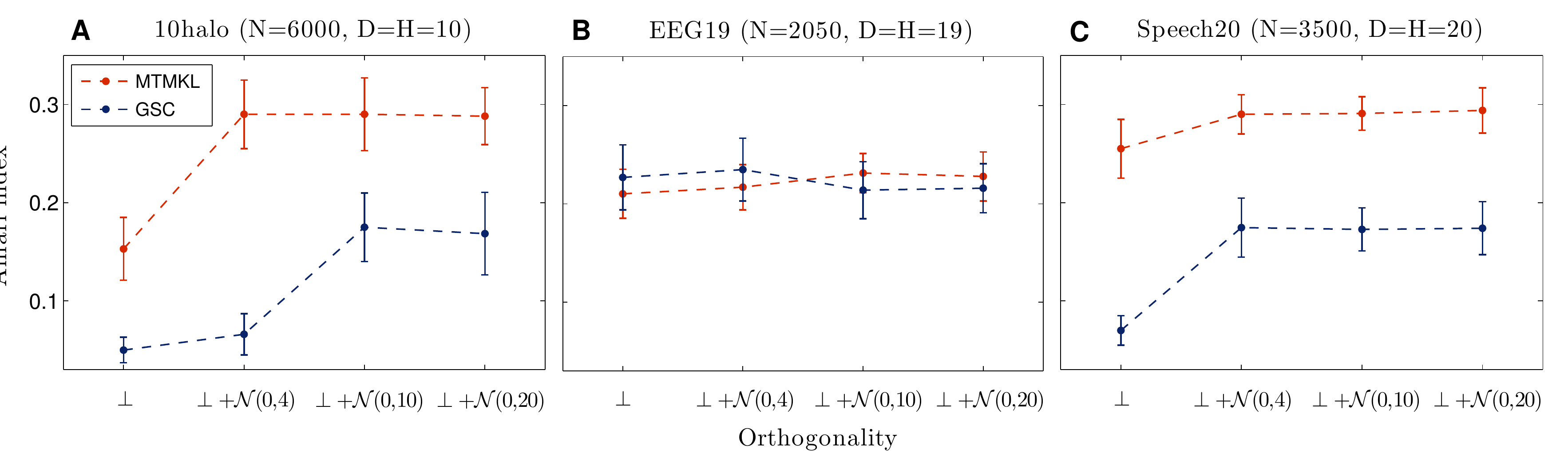}
\caption{Performance of GSC vs.\ MTMKL
on source separation benchmarks with varying degrees of orthogonality of the mixing
bases. The orthogonality on the x-axis 
varies from being orthogonal $\perp$ to increasingly non-orthogonal mixing as 
randomly generated orthogonal bases are perturbed by 
adding Gaussian noise $\NGauss(0,\sigma)$ to them. 
No observation noise was assumed for these experiments. 
Performances are compared on the Amari index \refp{EqnAmari}. }
\label{fig:gsc_vs_mtmkl_no_noise}
\end{center}
\end{figure}

\begin{figure}
\begin{center}
\includegraphics[width=.5\textwidth]{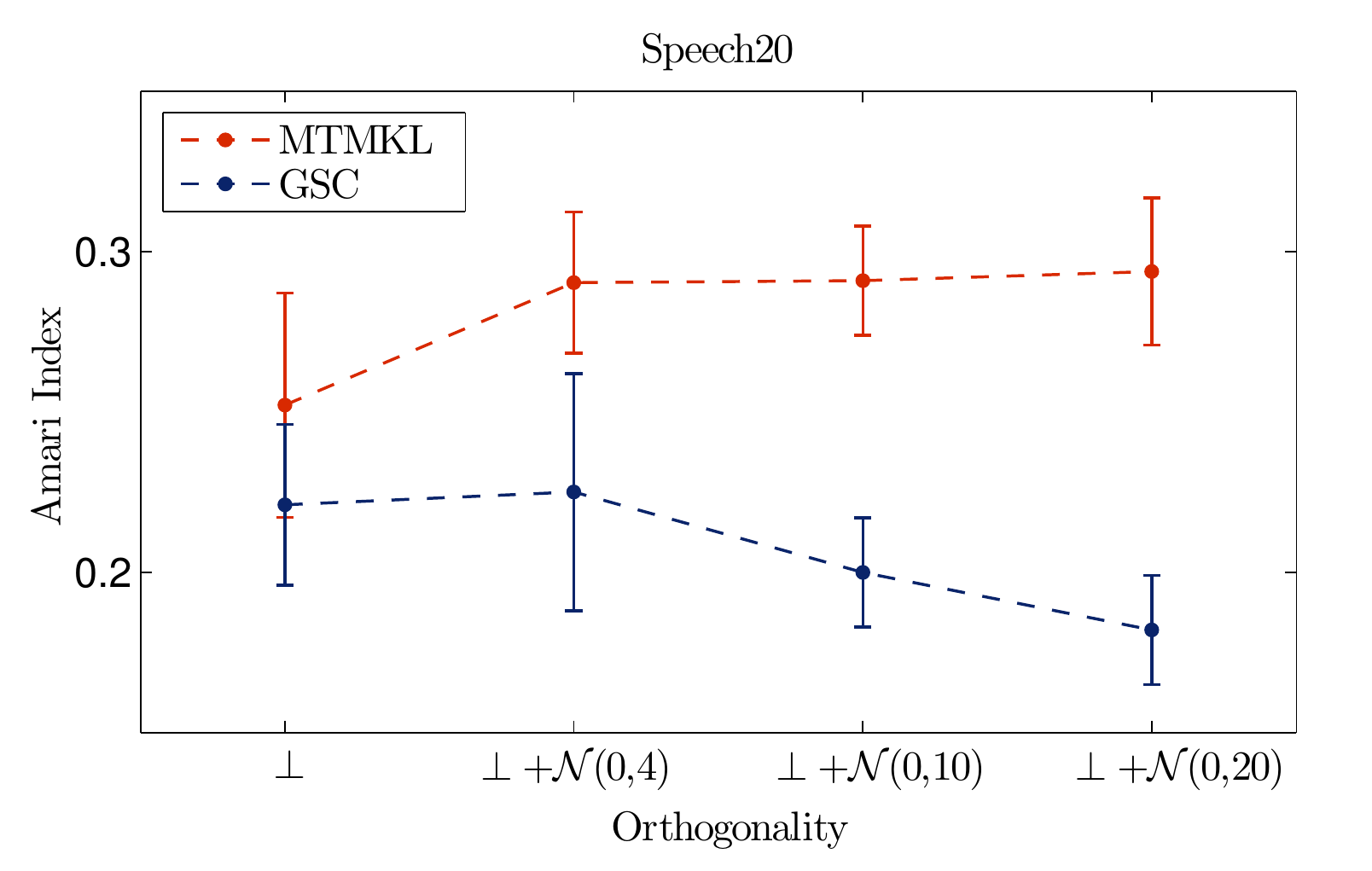}
\caption{Performance of GSC vs.\ MTMKL in terms of the Amari index \refp{EqnAmari} 
on the $\mathrm{Speech20}$ benchmark with varying degrees of orthogonality of the mixing
bases and Gaussian noise (with $\sigma=2$) added to observed data. The orthogonality on the x-axis 
varies from being orthogonal $\perp$ to increasingly non-orthogonal mixing as 
randomly generated orthogonal bases are perturbed by 
adding Gaussian noise $\NGauss(0,\sigma)$ to them.}
\label{fig:gsc_vs_mtmkl_Speech20_noisy}
\end{center}
\end{figure}

On synthetic data we have seen 
that spike-and-slab sparse coding can effectively recover 
sparse directions such as those generated by standard
sparse coding models. As many signals such as acoustic speech data are sparse, and as different
sources mix linearly, the assumptions of sparse coding match such data well. Source separation is
consequently a natural application domain of sparse coding approaches, and well suited for benchmarking
novel spike-and-slab as well as other sparse coding algorithms.
To systematically study the a-posteriori independence assumption in
factored variational approaches, we monitor the recovery of sparse
directions of GSC and MTMKL for an increasing degree of the mixing
matrix's non-orthogonality.
Figure \ref{fig:gsc_vs_mtmkl_no_noise} shows the performance of both the methods based on three different source separation
benchmarks obtained from \citep[ICALAB;][]{icalab2007}. The error bars show two standard deviations estimated based on
$15$ trials per experiment.
The x-axis in the figure represents the degree of orthogonality
of the ground truth mixing bases $W^{\mathrm{gen}}$. Starting from strictly orthogonal at the left, 
the bases were made increasingly non-orthogonal by randomly generating orthogonal bases and 
adding Gaussian distributed noise to them with $\sigma \in \{4,10,20\}$, respectively. For Figure \ref{fig:gsc_vs_mtmkl_no_noise} no observation noise was added to the mixed sources. 
For both the algorithms we performed 
$100$ iterations per run.\footnote{For the MTMKL algorithm we observed convergence after $100$ iterations while the GSC 
algorithm continued to improve with more iterations. However, allowing the same number of iterations to 
both the algorithms, the reported results are obtained with $100$ iterations.} The GSC truncation 
parameters $H^{\prime}$ and $\gamma$ were set to $10$ for all the following experiments, therefore for {\em 10halo} the GSC inference was exact. 
As can be observed, both approaches recover the sparse directions well. While performance on the EEG19 data set is
the same, GSC consistently performs better than MTMKL on {\em 10halo} and {\em Speech20}. 
If observation noise is added, the difference can become still more pronounced
for some data sets. Figure \ref{fig:gsc_vs_mtmkl_Speech20_noisy} shows the performance
in the case of $Speech20$ (with added Gaussian noise with $\sigma=2.0$), for instance.
Along the x-axis orthogonality decreases, again. 
While the performance of MTMKL decreases with decreasing orthogonality, 
performance of GSC increases in this case. For other data sets increased observation noise 
may not have such effects, however (see Appendix, Figure \ref{fig:gsc_vs_mtmkl_10halo_EEG19_noisy} for two examples).

Next we look at MAP based sparse coding algorithms 
for the source separation task. Publicly available methods which we compare with are
\citep[SPAMS;][]{MairalBPS09} 
and the efficient sparse coding algorithms \citep[ESCA;][]{LeeEtAl2007}.
These methods are based on linear regression with lasso regularization, where sparsity is induced by 
introducing a parameter-regulated penalty term in the objective function,\footnote{For both the 
algorithms compared here, optimal values for sparsity controlling regularization
parameters were chosen through cross-validation.} which penalizes the 
$L_1-$norm of regressors (or latent variables). In a probabilistic context this is equivalent to 
assuming a Laplace prior on the regressors. In this experiment we test the performance on another set of ICALAB \citep{icalab2007} 
benchmarks used previously \citep{SuzukiSugiyama2011,LuckeSheikh2012}. Following \cite{SuzukiSugiyama2011} we use $N=200$ and $N=500$ 
data points from each benchmark and generate observed data by mixing 
the benchmark sources with randomly generated orthogonal bases and adding no noise to the observed data.
For each experiment we performed $50$ trials with a new randomly generated orthogonal
data mixing matrix $W^{\mathrm{gen}}$ and new parameter initialization in each trial. The 
GSC inference was exact for these experiments with better results obtained with observed noise constrained 
to be homoscedastic. 
We performed up to $350$ iterations of the GSC algorithm (with more iterations continuing to improve the performance) while for the 
other algorithms we observed convergence between $100$ and $300$ iterations.

Table \ref{tbl:perf-src-sep} lists the performances of the algorithms.
As can be observed, the spike-and-slab based models perform better than
the standard sparse coding models for all except of one experiment (Sergio7,
$200$ data points) where SPAMS performs comparably well (or slightly better).
Among the spike-and-slab models, GSC performs best for all settings with $500$
data points, while MTMKL is better in two cases for $200$ data points.\footnote{In 
Table \ref{tbl:perf-src-sep} the results do not necessarily improve with an 
increased number of data points. However, the data 
points considered here are not independent samples. Following \citet[][]{SuzukiSugiyama2011} 
we always took consecutive $200$ or $500$ data points (after an offset) from each  
of the benchmarks. Therefore, due to time-dependencies in the signals, 
the underlying data point statistics change with the number of data points.}
Further improvements on some settings in
Table \ref{tbl:perf-src-sep} can be obtained by algorithms constrained
to assume orthogonal bases
\citep{SuzukiSugiyama2011,LuckeSheikh2012}.
However, for {\em 10halo} and {\em speech4} GSC and MTMKL are better without such
an explicit constraint.

\begin{table}
\begin{center}
\renewcommand{\arraystretch}{1.1}
\begin{center}
\begin{tabular}{|c|c|c|cccc|}\hline
\multicolumn{3}{|c|}{data sets} & \multicolumn{4}{c|}{Amari index - mean (std.)}\\\hline
name     & \hspace{-.05cm}H = D \hspace{-.05cm} & \hspace{-.3cm}N\hspace{-.3cm} & \hspace{-.4cm}GSC &\hspace{-.4cm} MTMKL  &\hspace{-.4cm} SPAMS & \hspace{-.4cm}ESCA\\\hline
\hspace{-.2cm} 10halo   & 10 & 200 & \hspace{-.2cm} 0.27(.04)      & \hspace{-.37cm} \bf{0.21(.05)} & \hspace{-.37cm}  0.28(0)      &\hspace{-.37cm} 0.31(.02)\\
                        & & 500    & \hspace{-.2cm} \bf{0.17(.03)} & \hspace{-.37cm} 0.20(.03)      & \hspace{-.37cm}  0.29(0)      &\hspace{-.37cm} 0.29(.02)\\\hline
\hspace{-.2cm} Sergio7  & 7 & 200  & \hspace{-.2cm}\bf{0.19(.05)}  & \hspace{-.37cm} \bf{0.19(.03)} & \hspace{-.37cm}  \bf{0.18(0)} &\hspace{-.37cm} 0.27(.04)\\
                        & & 500    & \hspace{-.2cm}\bf{0.13(.04)}  & \hspace{-.37cm} 0.23(.04)      & \hspace{-.37cm}  0.19(0)      &\hspace{-.37cm} 0.18(.04)\\\hline
\hspace{-.2cm} Speech4  & 4 & 200  & \hspace{-.2cm} \bf{0.13(.04)} & \hspace{-.37cm} \bf{0.14(.03)} & \hspace{-.37cm}  0.18(0)      &\hspace{-.37cm} 0.23(.02)\\
                          &  & 500 & \hspace{-.2cm}\bf{0.10(.04)}  & \hspace{-.37cm} 0.14(.08)      & \hspace{-.37cm}  0.16(0)      &\hspace{-.37cm} 0.17(0)\\\hline
\hspace{-.2cm} c5signals & 5 & 200 & \hspace{-.2cm} 0.29(.08)      & \hspace{-.37cm} \bf{0.24(.08)} & \hspace{-.37cm}  0.39(0)      &\hspace{-.37cm} 0.47(.05)\\
                          & & 500  & \hspace{-.2cm}\bf{0.31(.06)}  & \hspace{-.37cm} \bf{0.32(.03)} & \hspace{-.37cm}  0.42(0)      &\hspace{-.37cm} 0.48(.05)\\\hline
\end{tabular}
\end{center}
\caption{Performance of GSC, MTMKL and other publicly available
 sparse coding algorithms on benchmarks for source separation. 
Performances are compared based on the Amari index \refp{EqnAmari}. Bold values 
highlight the best performing algorithm(s).}
\label{tbl:perf-src-sep}
\end{center}
\end{table}

Figure~\ref{fig:sparse-post} was generated in a similar fashion on the {\em 10halo} data set. 
There we computed the exact posterior \refp{EqnPostS} over $H = 10$ latent dimensions, thus the approximation  
parameters were $\gamma = H'= H$ (exact E-step). After performing $50$ EM iterations and learning all the model parameters, 
we then visualized marginalized posteriors for a given data point along each column of the figure. 
The top row of the figure allows us get an idea of how concentrated and sparse a data point is in terms of 
the latents contributing to its posterior mass. 
The bottom row of the figure on the other hand allows us to observe the sparsity in the posterior w.r.t.\ the dimensionality of 
the hyperplanes spanned by the latents, with a posterior mass accumulation in low-dimensional hyperplanes.

\subsection{Computational Complexity vs.\ Performance}
\label{sec:CompvsPerf}
\begin{figure}
\begin{center}
\begin{minipage}[b]{5.5cm}
\includegraphics[width=\textwidth]{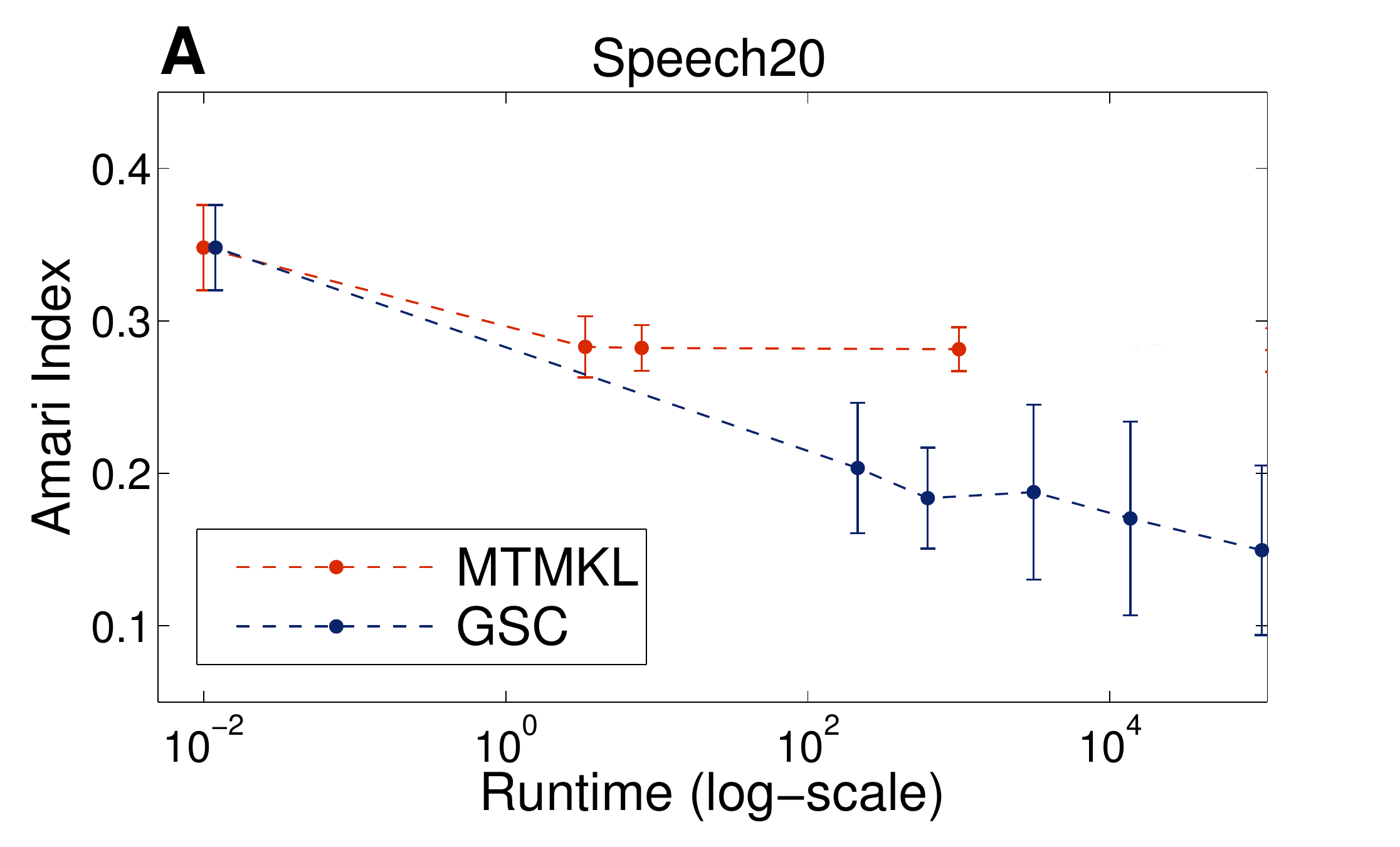}
\end{minipage}
\raisebox{1.58cm}{
\hspace{-.7cm}
\begin{minipage}{5.5cm}
\includegraphics[width=\textwidth]{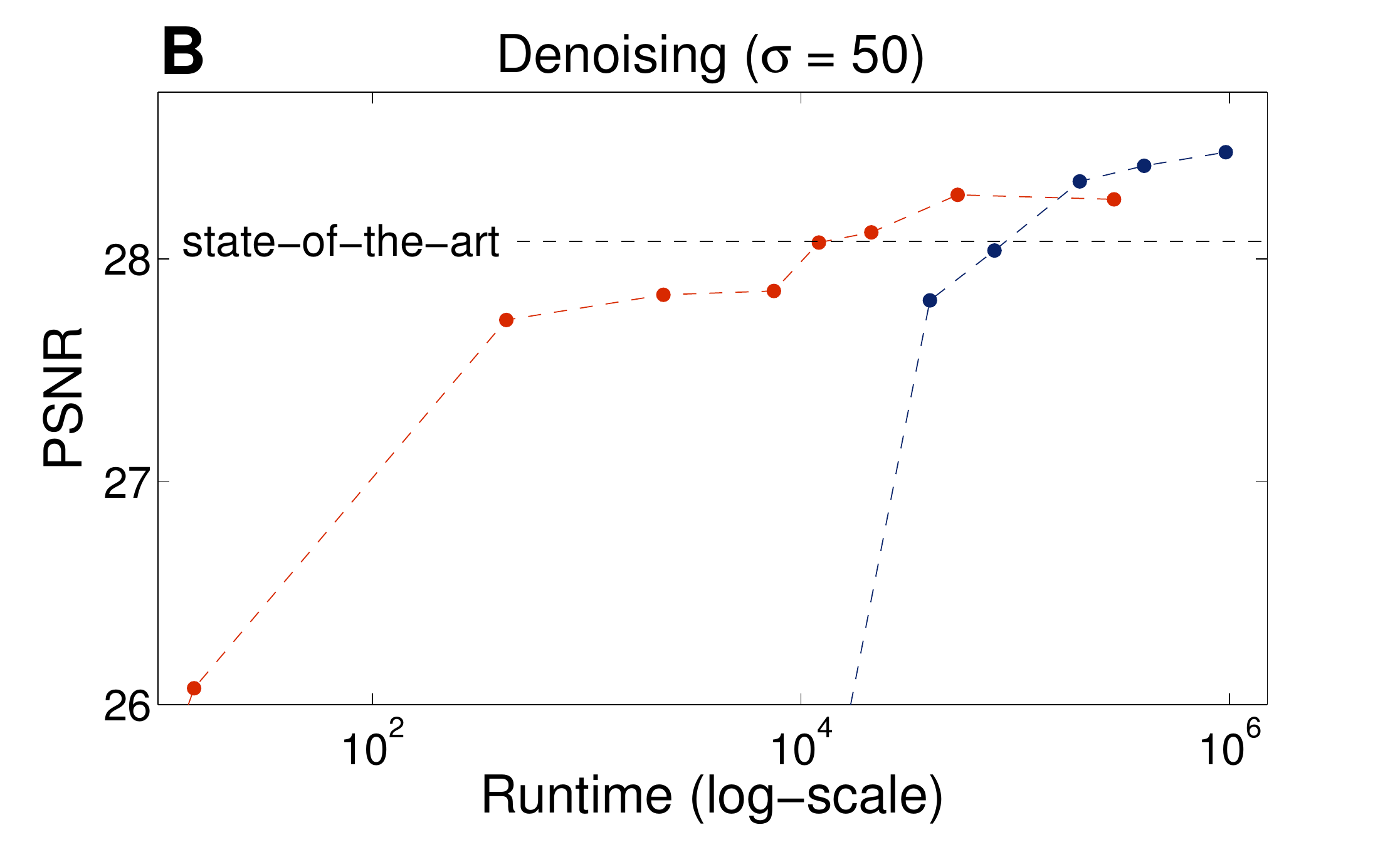}
\end{minipage} 
\raisebox{0.04cm}{
\hspace{-.7cm}
\begin{minipage}{5.5cm}
\includegraphics[width=.99\textwidth]{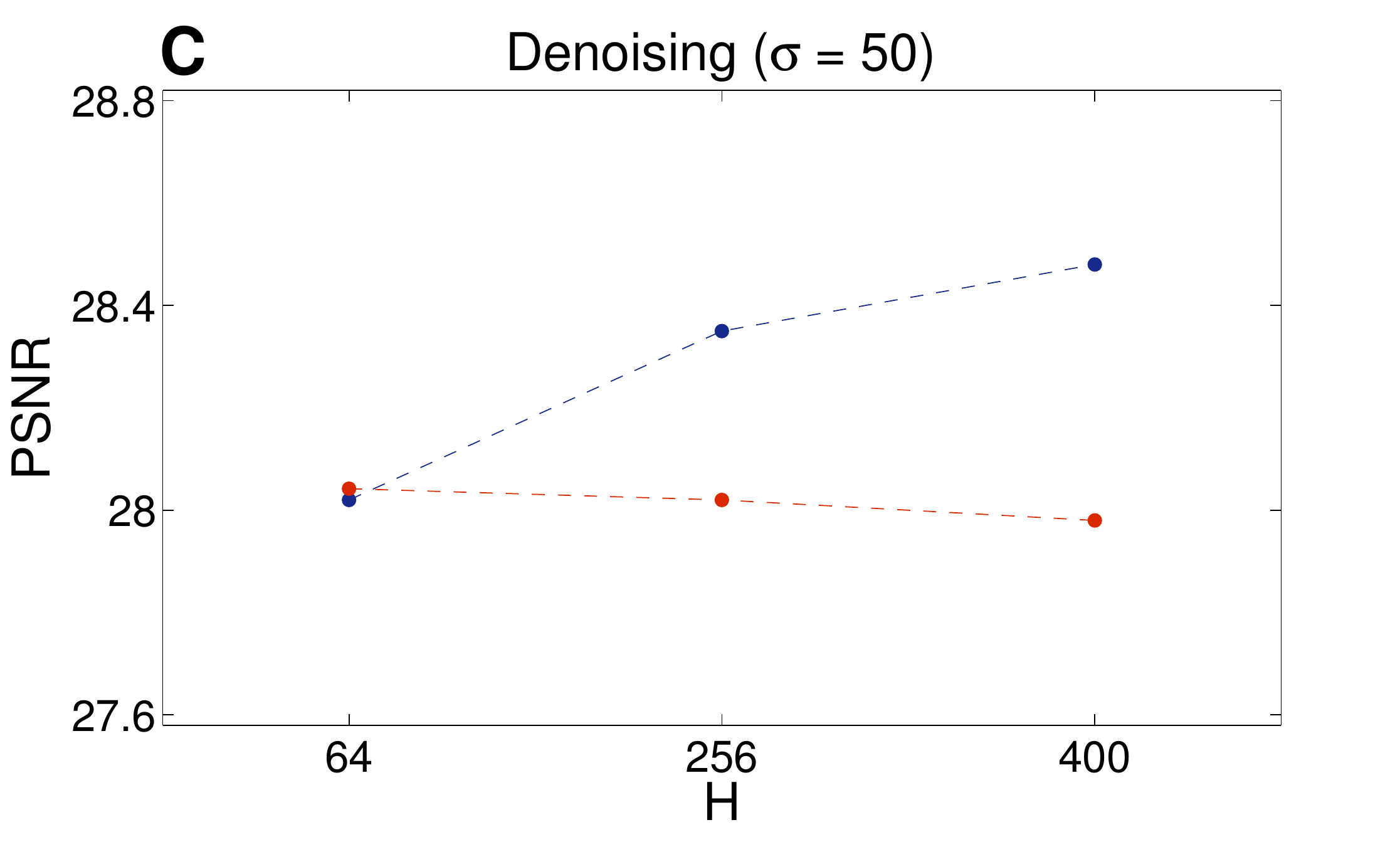}
\end{minipage}
}
}
\caption{{\bf A,B:} Runtime vs.\ performance comparison of GSC (blue) and MTMKL (red) 
on source separation and denoising tasks. Source separation is compared on the Amari index (the lower the better)
while the denoising is compared on the peak signal-to-noise (PSNR) ratio (the higher the better).
{\bf C}: Performance of GSC (blue) and MTMKL (red) on the denoising task against an increasing 
number of latents.}
\label{fig:gsc_vs_mtmkl_runtime}
\end{center}
\end{figure}

In terms of computational complexity, GSC and MTMKL algorithms are significantly different,
so we also looked at the trade-off between their computational costs
versus performance. Subfigures A and B in Figure \ref{fig:gsc_vs_mtmkl_runtime} show
performance against compute time for both algorithms. The error bars for the $\mathrm{Speech20}$ plot
were generated from $15$ trials per experiment. For MTMKL we obtained the 
plot by increasing the number of iterations from $50$ to $100$ and $1000$, while for 
the GSC plot we performed $100$ iterations with 
$H^{\prime} = \gamma \in [2,3,5,7,10]$. 
For the image denoising task (described next), the MTMKL plot was generated from a run with $H=64$ 
latents
and the number of iterations going up to $12$K. The GSC plot was 
generated from $H=400$ latents with $H^{\prime}$ and $\gamma$ being $10$ and $5$ respectively. 
The last point on the GSC (blue) curve corresponds to the $120th$ EM iteration. As can
be observed for both tasks, the performance of MTMKL saturates from
certain runtime values onwards. GSC on the other hand continues to show improved
performance with increasing computational resources.

For the denoising task we also compared the performance of both the algorithms 
against an increasing number of latents $H$. While the computational cost of the MTMKL 
algorithm increases linearly w.r.t.\ $H$, the runtime cost of the truncated variational 
GSC remains virtually unaffected by it, since it scales w.r.t.\ the parameters $\HPrime$ 
and $\gamma$ (see Section \ref{sec:ET-complexity}). 
In this experiment we performed $65$ iterations of the GSC algorithm for $H \in \{64,256\}$ and up to 
$120$ iterations for $H=400$. For MTMKL we performed up to 
$120$ iterations for each given $H$. Figure \ref{fig:gsc_vs_mtmkl_runtime}C
summarizes the results of this experiment. In the figure we can see a constant performance increase for 
GSC, while for MTMKL we actually observe a slight decrease in performance. 
This is in conformity with what \cite{TitsiasLazaro2011}  
report in their work that for the denoising task they observed no performance improvements for larger number of latents.

\subsection{Image Denoising}  
\begin{figure}
\begin{center}
\vspace{3mm}
\begin{minipage}[b]{10cm}
    \begin{minipage}[b]{5cm}
    \includegraphics[width=.98\textwidth]{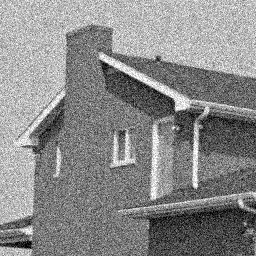}
    \end{minipage}
  \raisebox{2.35cm}{
    \begin{minipage}{5cm}
    \includegraphics[width=.98\textwidth]{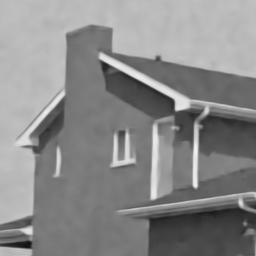}
    \end{minipage} }
    \label{fig:house-denoising-25}
\end{minipage}
\vspace{2mm}
\includegraphics[width=.7\textwidth]{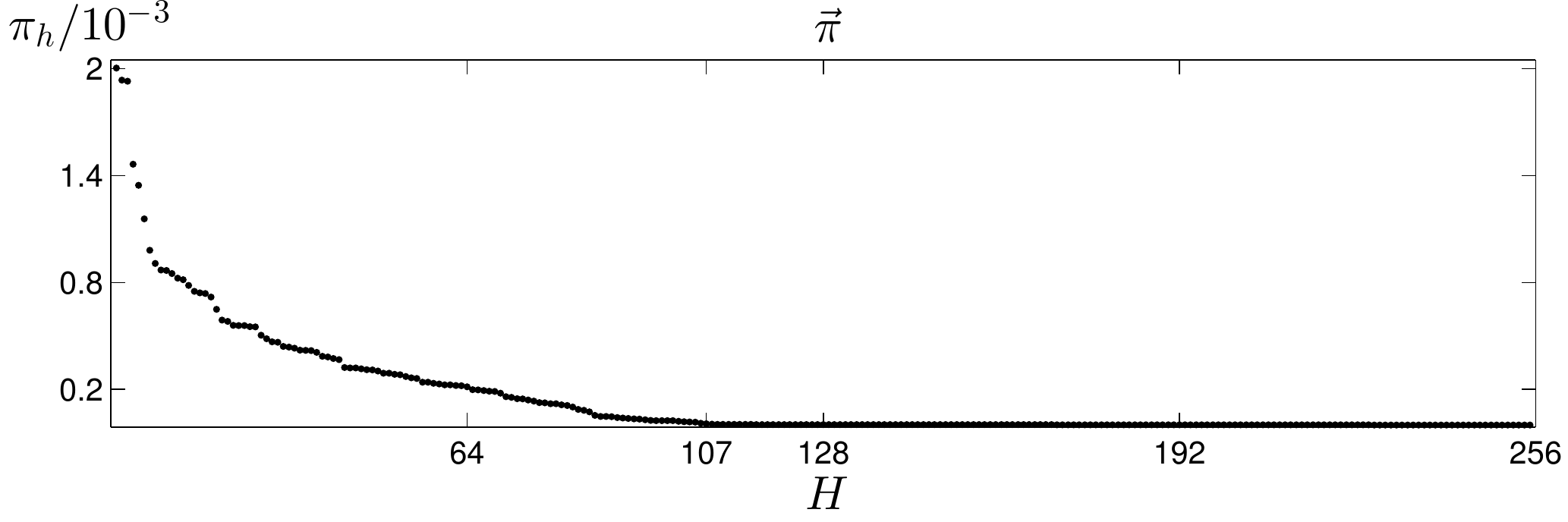}
\includegraphics[width=.72\textwidth]{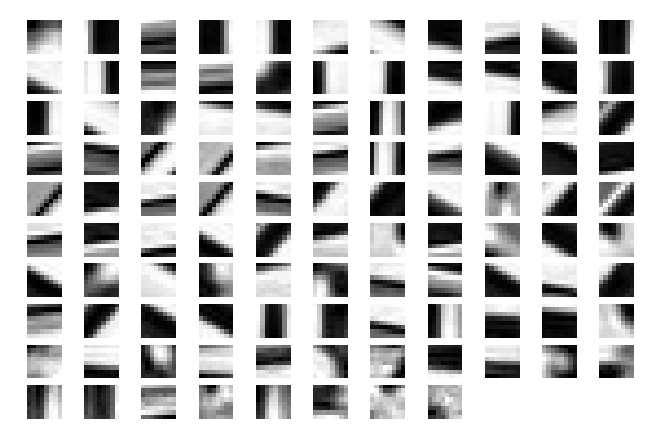}
\caption{Top left: Noisy ``house'' image with $\sigma = 25$. Top right: GSC denoised
image. Middle: Inferred sparsity values $\pi_h$ in descending order indicate  
that finally around $107$ of in total $256$ latent dimensions significantly contribute
to model the data. Bottom: Basis functions (ordered from left to right, top to bottom) 
corresponding to the first $107$ latent dimensions sorted w.r.t.\ the decreasing
sparsity values $\pi_h$ .}
\label{fig:denoising-25}
\end{center}
\end{figure}

Finally, we investigate performance of the GSC
algorithm on the standard ``house'' benchmark for denoising which has been used
for the evaluation of similar approaches \citep[e.g.,][]{LiLiu2009,ZhouetAl2009} including the MTMKL
spike-and-slab approach. The MTMKL approach currently represents the 
state-of-the-art on this benchmark \citep[see][]{TitsiasLazaro2011}.
We also compare with the approach by \citet[][]{ZhouetAl2009} as a representative sampling-based optimization scheme.
For the task a noisy input image is generated by adding Gaussian noise (with zero mean and standard deviation 
determining the noise level) to the $256\times256$ image (see Figure \ref{fig:denoising-25}). 
Following the previous studies, we generated
$62,001$ overlapping (shifted by $1$ pixel) $8$ $\times$ $8$ patches from the noisy image.
We then applied $65$ iterations of the GSC algorithm for $H \in \{64,256\}$ 
for different noise levels $\sig \in \{15,25,50\}$. The truncation parameters $H^{\prime}$ and $\gamma$ 
for each run are listed in Table \ref{tbl:denoising}. We assumed homoscedastic observed noise with
a priori unknown variance in all these experiments (as the MTMKL model). 

A comprehensive comparison of the denoising results of the various algorithms is shown in 
Table \ref{tbl:denoising}, where performance is measured in terms of the peak signal-to-noise (PSNR) ratio. 
We found that for the low noise level ($\sigma=15$) GSC
is competitive with other approaches but with MTMKL performing slightly better.
For the higher noise levels of $\sigma=25$ and $\sigma=50$, GSC outperforms all the 
other approaches including the MTMKL approach that represented the state-of-the-art. 
In Figure \ref{fig:denoising-25} we show our result for noise level $\sigma=25$. The figure contains 
both the noisy and the GSC denoised image along with the
inferred sparsity vector $\piVec$ and all bases with appearance probabilities
significantly larger than zero (sorted from high such probabilities to low ones).
We also applied GSC with higher numbers of latent dimensions: 
Although for low noise levels of $\sigma=15$ and $\sigma=25$ we did not measure significant improvements, 
we observed a further increase for $\sigma=50$. 
For instance, with $H=400$, $H'=10$ and $\gamma=8$, we obtained for $\sigma=50$ a PSNR of $28.48\mathrm{dB}$. 

As for source separation described in Section \ref{sec:CompvsPerf}, we also 
compared performance vs.\ computational demand of both algorithms for the task of image denoising. 
As illustrated in A and B of Figure \ref{fig:gsc_vs_mtmkl_runtime}, 
MTMKL performs better when computational resources are relatively limited. However, when increasingly more computational
resources are made available, MTMKL does not improve much further on its performance while
GSC performance continuously increases and eventually outperforms MTMKL on this task. 

\newcommand{\vertlarge}{\phantom{$\disS\int^f$}}
\newcommand{\vertsmall}{\phantom{\hspace{-2mm}$\disS\sum$\hspace{-2mm}}}
\begin{table}[h]
\footnotesize
\scriptsize
\begin{center}
\begin{tabular}{|c|c|c|cc|c|cc|}\hline
 \hspace{-.5cm} & \multicolumn{7}{c|}{\vertlarge PSNR (dB)\vertlarge }\\\cline{2-8}
\hspace{-.2cm} Noise \hspace{-.2cm} & \hspace{-.1cm}Noisy img \hspace{-.2cm} & MTMKL$^{exp.}$ \hspace{-.3cm} & K-SVD$^{mis.}$\hspace{-.8cm}&\hspace{-.25cm}*K-SVD$^{match}$ \hspace{-.3cm}  &\hspace{-.2cm} Beta pr. \hspace{-.2cm} & \vertlarge \hspace{-.45cm}GSC \tiny{(H=64)}\hspace{-.2cm}&\hspace{-.1cm} GSC \tiny{(H=256)}\\\hline
$\sigma$=15   &  \vertsmall 24.59\vertsmall   &  \bf{34.29}  & \hspace{-.22cm} 30.67\hspace{-.22cm}&  \hspace{-.22cm} 34.22\hspace{-.22cm}&  34.19 & 32.68 \tiny{(H'=10,$\gamma$=8)} & 33.78 \tiny{(H'=18,$\gamma$=3)}  \\\hline
$\sigma$=25  &   \vertsmall 20.22\vertsmall   &  31.88  & \hspace{-.22cm} 31.52\hspace{-.22cm}& \hspace{-.22cm} 32.08\hspace{-.22cm}&  31.89 & 31.10 \tiny{(H'=10,$\gamma$=8)} & {\bf{32.01}} \tiny{(H'=18,$\gamma$=3)}   \\\hline
$\sigma$=50 &   \vertsmall 14.59\vertsmall    &  28.08  & \hspace{-.22cm} 19.60\hspace{-.22cm}& \hspace{-.22cm} 27.07\hspace{-.22cm}&  27.85 & 28.02 \tiny{(H'=10,$\gamma$=8)} & {\bf{28.35}} \tiny{(H'=10,$\gamma$=8)}  \\\hline
\end{tabular}
\end{center}
\caption{Comparison of the GSC algorithm with other methods applied to the ``house'' benchmark.
The compared methods are: MTMKL~\citep{TitsiasLazaro2011}, K-SVD~\citep{LiLiu2009}, and 
Beta process~\citep{ZhouetAl2009}. Bold values 
highlight the best performing algorithm(s). $^\ast$High values for K-SVD matched are not made bold-faced as
the method assumes the noise variance to be known a-priori \citep[see][]{LiLiu2009}.
}
\label{tbl:denoising}
\end{table}

\section{Discussion}
\label{Sec:Discussion}
The last years have seen a surge in the application of sparse coding
algorithms to different research domains, along with developments
of new sparse coding approaches with increased capabilities. 
There are currently different lines of research followed for
developing new algorithms: one direction is based on the standard sparse
coding algorithm \citep{OlshausenField1996} with Laplace prior and
parameter optimization using maximum a-posteriori (MAP) estimates of
the latent posteriors for efficient training. This original approach has since been
made more efficient and precise. Many sparse coding algorithms based on the MAP estimation are
continuously being developed and are successfully applied in a variety of 
settings \citep[e.g.,][]{LeeEtAl2007,MairalBPS09}.
Another line of research aims at a fully Bayesian description of
sparse coding and emphasizes greater flexibility by using different
(possibly non-Gaussian) noise models and estimations of the number of
hidden dimensions. The great challenge of these general models is the
procedure of parameter estimation. For instance, the model by
\citet[][]{MohamedEtAl2012} uses Bayesian methodology
involving conjugate priors and hyper-parameters in combination with
Laplace approximation and different sampling schemes.

A line of research falling in between conventional and fully
Bayesian approaches is represented by the truncated variational approach studied here
and by other very recent developments \citep{TitsiasLazaro2011,GoodfellowEtAl2013}.
While these approaches are all based on spike-and-slab
generalizations of sparse coding (like fully Bayesian approaches), they
maintain deterministic approximation procedures for parameter
optimization. Variational
approximations allow for applications to large hidden spaces which pose a challenge 
for sampling approaches especially in cases of multi-modal posteriors.
Using the novel and existing approaches in different experiments of this study,
we have confirmed the advantages of spike-and-slab priors for
sparse coding, and the scalability of variational approximations for such models. 
The newly developed truncated variational algorithm scales almost linearly with the number of
hidden dimensions for fixed truncation parameters (see for instance the scaling behavior in supplemental Figure \ref{fig:complexity} for $H$ going up to $1024$). 
The MTMKL algorithm by \cite{TitsiasLazaro2011} has been applied on the same scale.
Using a similar approach also based on factored distributions
\citet{GoodfellowEtAl2013} report results for up to a couple of thousands
latent dimensions (albeit on small input dimensions and having a more constrained
generative model).
Sampling based algorithms for non-parametric and fully Bayesian
approaches are more general but have not been applied to such large scales.

A main focus of this work and reasoning behind the algorithm's development 
is due to the long-known biases introduced by factored
variational approximations
\citep{MacKay2001,IlinValpola2005,TurnerSahani2011}. Our systematic
comparison of the GSC algorithm to the method by
\cite{TitsiasLazaro2011} 
confirms the earlier observation
\citep{IlinValpola2005} that factored variational approaches are
biased towards orthogonal bases. If we compare the performance of both
algorithms on the recovery of non-orthogonal sparse directions, the
performance of the factored variational approach is consistently lower
than the performance of the truncated variational algorithm 
(Figure \ref{fig:GSC-ET_vs_MTMKL_non-ortho}). 
The same applies for experiments
for unmixing real signals in which we increased the non-orthogonality (Figure \ref{fig:gsc_vs_mtmkl_no_noise}A,C; suppl.\,Figure \ref{fig:gsc_vs_mtmkl_10halo_EEG19_noisy});
although for some data performance is very similar (Figure \ref{fig:gsc_vs_mtmkl_no_noise}B).
Also if sources are mixed orthogonally, we usually observe better
performance of the truncated variational approach
(Table \ref{tbl:perf-src-sep}), which is presumably due to the more general
underlying prior (i.e., a fully parameterized Gaussian slab). 
Overall, GSC is the best performing algorithm on source separation tasks 
involving non-orthogonal sparse directions \citep[compare][for algorithms constrained
to orthogonal bases]{SuzukiSugiyama2011}.
For some data sets with few data points, we observed an equal
or better performance of the MTMKL approach, which can be explained 
by their Bayesian treatment of the model parameters
(see Table \ref{tbl:perf-src-sep}, performance with $200$ data points).
Notably, both approaches are consistently better on source separation
benchmarks than the standard sparse coding approaches SPAMS
\citep{MairalBPS09} and ESCA \citep{LeeEtAl2007} (see
Table \ref{tbl:perf-src-sep}).  This may be taken as evidence for the
better suitability of a spike-and-slab prior for such types of data. 

For source separation our approach (like conventional sparse coding or ICA) seeks to infer  
sparse directions by capturing the sparse, latent structures from the spatial domain of the input signals. 
However, when dealing with data that also carry a temporal structure (e.g., speech or EEG recordings),
other approaches which explicitly model temporal regularities 
such as Hidden Markov Models (HMMs) may as well be a more natural and (depending on the task) a more 
suitable choice. Such methodologies can in principle be combined with the sparse coding approaches studied and
compared here to form more comprehensive models for spatio-temporal data, which can yield improved performance 
on blind source separation tasks \citep[compare e.g.,][]{VanEtAl2009,GauthamEtAl2010}.

In the last experiment of this study, we finally compared the performance of factored
and truncated variational approximations on a standard image denoising task (see
Table \ref{tbl:denoising}).  The high PSNR values observed for both
approaches again in general speak for the strengths of spike-and-slab
sparse coding. The MTMKL model represented the state-of-the-art on
this benchmark, so far. Differences of MTMKL to previous approaches
are small, but this is due to the nature of such
long-standing benchmarks (compare, e.g., the MNIST data set).  For the
same denoising task with standard noise levels of $\sigma=25$ and
$\sigma=50$ we found the GSC model to further improve the
state-of-the-art (compare Table \ref{tbl:denoising} with data by
\citealtt{LiLiu2009}, \citealtt{ZhouetAl2009}, \citealtt{TitsiasLazaro2011}).
While we observed a continuous increase of performance with the number of hidden
dimensions used for GSC, the MTMKL algorithm \citep{TitsiasLazaro2011}
is reported to reach saturation at $H=64$ latent dimensions.
As the learned sparse directions become less and less orthogonal the
more over-complete the setting gets, this saturation may again be due
to the bias introduced by the factored approach. GSC with $H=256$
improves the state-of-the-art with $32.01$dB for $\sigma=25$ and with
$28.35$dB for $\sigma=50$ (with even higher PSNR for $H=400$). As we
assume an independent Bernoulli prior per latent dimension, GSC can 
also prune out latent dimensions by inferring very low
values of $\pi_h$ for the bases that make negligible contribution in 
the inference procedure. This can be observed in Figure \ref{fig:denoising-25}, 
where for the application of GSC to
the denoising task with $\sigma=25$, we found only about
$107$ of the $256$ basis functions to have significant probabilities
to contribute to the task. This means that GSC with about
$100$ basis functions can be expected to achieve almost the same
performance as GSC with $256$ basis functions. However, in practice we
observed that the average performance increases with more
basis functions because local optima can more efficiently be avoided.
This observation is not limited to the particular approach
studied here; also for other approaches to sparse learning, efficient
avoidance of local optima has been reported if the number of assumed
hidden dimensions was increased \citep[e.g.][]{Spratling2006,LuckeSahani2008}.
In comparison to MTMKL, GSC can make use of significantly more basis functions.
It uses about $100$ functions while MTMKL performance saturates at about $64$
as mentioned previously. On the other hand, we found MTMKL to perform
better on the low noise level setting (see $\sigma=15$ in
Table \ref{tbl:denoising}) or when relatively limited
computational resources are available (see
Figure \ref{fig:gsc_vs_mtmkl_runtime}).

In conclusion, we have studied a novel learning algorithm for sparse
coding with spike-and-slab prior and compared it with a number of
sparse coding approaches including other spike-and-slab based methods.
The results we obtained show that the truncated EM approach is
a competitive method. It shows that posterior dependencies and
multi-modality can be captured by a scalable deterministic approximation. 
Furthermore, the direct comparison with a factored variational approach in source separation
experiments confirms earlier observations that
assumptions of a-posteriori independence introduces biases, and that
avoiding such biases, e.g.\ by a truncated approach, improves the
state-of-the-art on source separation benchmarks as well as on
standard denoising tasks. However, we also find that under certain
constraints and settings, factored variational learning for
spike-and-slab sparse coding may perform as well or better.
In general, our results argue in favor of spike-and-slab
sparse coding models and recent efforts for developing improved algorithms
for inference and learning in such models.

\subsection*{Acknowledgements}
We acknowledge funding by the German Research Foundation (DFG), grant LU 1196/4-2,
and by the German Ministry of Research and Education (BMBF), grant 01GQ0840 (BFNT Frankfurt). 
Furthermore, we acknowledge support by the Frankfurt Center for Scientific Computing (CSC Frankfurt).

\renewcommand{\thesection}{\alph{section}}
\renewcommand{\thesubsection}{\Alph{section}.\arabic{subsection}}
\appendix
\section{Derivation of M-step Equations}
\label{sec:UpdRulesDerv}
Our goal is to optimize the free-energy w.r.t. $\Theta$:
\begin{eqnarray*}
\FF(\ThetaOld,\Theta) &=&
    \sum\limits_{n=1}^{N} \left\langle \log p(\yVecN, \sVec, \zVec\,|\,\Theta)
    \right\rangle_n + H(\ThetaOld)\\
 &=& \sum\limits_{n=1}^{N} \sum\limits_{\sVec}\int\limits_{\zVec}\ p(\sVec,\zVec\,|\,\yVecN,\ThetaOld) \nonumber\\
 & & \disT  \Big[\log\big( p(\yVecN\,|\,\sVec,\zVec,\Theta)\big)+ \log \big( p(\zVec\,|\sVec,\,\Theta) \big) 
 \disT  + \log \big( p(\sVec\,|\,\Theta) \big) \Big]\,\dz\,+\,H(\ThetaOld)\,,
\end{eqnarray*}
where 
\begin{eqnarray*}
\log\big( p(\yVecN\,|\,\sVec,\zVec,\Theta)\big) &=&
- \frac{1}{2} \left(\log(2\pi^D) + \log|\Sigmad|\right) \nonumber\\
& & - \frac{1}{2}\left( \yVecN\, - \,W(\sVec\odot\zVec) \right)^\TT\,
\Sigmad^{-1}\,
\left( \yVecN\, - \,W(\sVec\odot\zVec) \right), \nonumber \\
\log\big(p(\zVec\,|\sVec,\,\Theta)\big)  &=&
- \frac{1}{2} \left(\log(2\pi^{|\sVec|}) + \log|\Sigmah \odot \sVec\,\sVec^{\TT}|\right) \\
& & - \frac{1}{2}\big(\left(\zVec\, - \,\muVec \right)\odot \sVec\ \big)^\TT\,
\big(\Sigmah \odot \sVec\,\sVec^{\TT}\big)^{-1}\,
\big(\left(\zVec\, - \,\muVec \right)\odot \sVec\ \big)
\\
\mbox{and}\hspace{2.75cm}
\log\big(p(\sVec\,|\Theta)\big) 
&=&  \disS\sum_{h=1}^{H} \log\left(\pi_h^{s_h}\,(1-\pi_h)^{1-s_h} \right)
.
\end{eqnarray*}
The free-energy thus takes the form:
\begin{eqnarray*}
\FF(\ThetaOLD,\Theta)
 &=& \sum\limits_{n=1}^{N} \sum\limits_{\sVec}\int\limits_{\zVec}\ q_n(\sVec,\zVec;\ThetaOLD) \nonumber\\
 & & \disT  \Big[- \frac{1}{2} \left(\log(2\pi^D) + \log|\Sigmad|\right) 
 - \frac{1}{2}\left( \yVecN\, - \,W(\sVec\odot\zVec) \right)^\TT\,
\Sigmad^{-1}\,
\left( \yVecN\, - \,W(\sVec\odot\zVec) \right) \\
 & & \disT- \frac{1}{2} \left(\log(2\pi^{|\sVec|}) + \log|\Sigmah \odot \sVec\,\sVec^{\TT}|\right) \\
& & - \frac{1}{2}\big(\left(\zVec\, - \,\muVec \right)\odot \sVec\ \big)^\TT\,
\big(\Sigmah \odot \sVec\,\sVec^{\TT}\big)^{-1}\,
\big(\left(\zVec\, - \,\muVec \right)\odot \sVec\ \big)\\
& & \disT +  \disT\sum_{h=1}^{H} \log\left(\pi_h^{s_h}\,(1-\pi_h)^{1-s_h} \right)  \Big]\,\dz\,+\,H(\ThetaOld)\,,
\end{eqnarray*}
where $q_n(\sVec,\zVec;\ThetaOLD)$ denotes the posterior $p(\sVec,\zVec\,|\,\yVecN,\ThetaOld)$. Now we can 
derive the M-step equations \refp{EqnMStepStart} to \refp{EqnMStepEnd} by canonically setting the derivatives 
of the free-energy above w.r.t.\ each parameter in $\Theta$ to zero.

\subsection{Optimization of the Data Noise}
Let us start with the derivation of the M-step equation for $\Sigmad$:
\begin{equation*}
\begin{split}
\frac{\partial}{\partial \Sigmad}& \FF(\ThetaOLD,\Theta)\\
& = \sum\limits_{n=1}^{N} \sum\limits_{\sVec}\int\limits_{\zVec} q_n(\sVec,\zVec;\ThetaOLD) \\
& \phantom{iiiiiiiiii} 
\disT  \Big[- \frac{1}{2} \frac{\partial}{\partial \Sigmad} \left(\log|\Sigmad|\right) 
 - \frac{1}{2}\frac{\partial}{\partial \Sigmad}\left( \yVecN\, - \,W(\sVec\odot\zVec) \right)^\TT\,
\Sigmad^{-1}\,
\left( \yVecN\, - \,W(\sVec\odot\zVec) \right) \Big]\dz\\
& = \sum\limits_{n=1}^{N} \sum\limits_{\sVec}\int\limits_{\zVec} q_n(\sVec,\zVec;\ThetaOLD) \\
& \phantom{iiiiiiiiii} \disT  \Big[- \frac{1}{2} \Sigmad^{-1}
 + \frac{1}{2}\Sigmad^{-2} \left( \yVecN\, - \,W(\sVec\odot\zVec) \right)\,
\left( \yVecN\, - \,W(\sVec\odot\zVec) \right)^\TT\, \Big]\dz \overset{!}{=} 0 \\ \\
%
%
%
%
\Rightarrow  
\Sigmad &= \frac{1}{N} \sum\limits_{n=1}^{N} \sum\limits_{\sVec}\int\limits_{\zVec} q_n(\sVec,\zVec;\ThetaOLD) 
 \disT  \Big[ \left( \yVecN\, - \,W(\sVec\odot\zVec) \right)
\left( \yVecN\, - \,W(\sVec\odot\zVec) \right)^\TT\, \Big] \dz\\
& 
 = \frac{1}{N}\sum_{n=1}^{N}\Big[ \Big(\yVecN - W\E{(\sVec\odot\zVec)}_n\Big)
	\Big(\yVecN - W\E{(\sVec\odot\zVec)}_n\Big)^{\TT} \nonumber \\
    & \hspace{4.5cm} + W\big[\E{(\sVec\odot\zVec)(\sVec\odot\zVec)^{\TT}}_n - \E{(\sVec\odot\zVec)}_n\E{(\sVec\odot\zVec)}_n^{\TT}\big]W^{\TT}\Big]\\
&
  = \frac{1}{N}\sum_{n=1}^{N}\Big[ \yVecN(\yVecN)^{\TT}
        - W\big[\E{(\sVec\odot\zVec)}_n\E{(\sVec\odot\zVec)}_n^{\TT}\big]W^{\TT}\Big]
        ,
\end{split}
\end{equation*}
where $\langle \,\cdot\, \rangle_n$ denotes the expectation value in Equation \refp{EqnExpWRTPost}.

\subsection{Optimization of the Bases}

We will now derive the M-step update for the basis functions $W$:

\begin{align*}
\frac{\partial}{\partial W}& \FF(\ThetaOLD,\Theta) \\
& = \sum\limits_{n=1}^{N} \sum\limits_{\sVec}\int\limits_{\zVec} q_n(\sVec,\zVec;\ThetaOLD) 
 \disT  \Big[- \frac{1}{2}\frac{\partial}{\partial W}\left( \yVecN\, - \,W(\sVec\odot\zVec) \right)^\TT\,
\Sigmad^{-1}\,
\left( \yVecN\, - \,W(\sVec\odot\zVec) \right) \Big] \dz \\
& = \sum\limits_{n=1}^{N} \sum\limits_{\sVec}\int\limits_{\zVec} q_n(\sVec,\zVec;\ThetaOLD) 
 \disT  \Bigg[- \frac{1}{\Sigmad\,}\bigg(\yVecN(\sVec\odot\zVec)^{\TT}
        - W(\sVec\odot\zVec)(\sVec\odot\zVec)^{\TT} \bigg) \Bigg] \dz \overset{!}{=} 0 \\ \displaybreak[1]\\
%
%
%
%
\Rightarrow W &  = \frac{\sum_{n=1}^{N} \yVecN \E{\sVec\odot\zVec}^{\TT}_n}
  {\sum_{n=1}^{N} \E{(\sVec\odot\zVec)(\sVec\odot\zVec)^{\TT}}_n }.
\end{align*}

\subsection{Optimization of the Sparsity Parameter}

Here we take the derivative of the free-energy w.r.t. $\piVec$:
\begin{align*}
\frac{\partial}{\partial \piVec}\FF(\ThetaOLD,\Theta) 
& = \sum\limits_{n=1}^{N} \sum\limits_{\sVec}\int\limits_{\zVec} q_n(\sVec,\zVec;\ThetaOLD) 
 \disT  \Bigg[\frac{\partial}{\partial \piVec} 
  \Big( \sVec\,\log \piVec + (1 -\sVec)\,log(1-\piVec) \Big) \Bigg] \dz \\
& = \sum\limits_{n=1}^{N} \sum\limits_{\sVec} q_n(\sVec;\ThetaOLD) 
 \disT  \Big[ \frac{\sVec}{\piVec} - \frac{(1-\sVec)}{(1-\piVec)}\Big] \overset{!}{=} 0 \\ \\
%
%
%
%
\Rightarrow  
\piVec & = \frac{1}{N}\sum_{n=1}^{N} \E{\sVec}_n.
\end{align*}

\subsection{Optimization of the Latent Mean}

Now we derive the M-step update for the mean $\muVec$ of the Gaussian slab:
\begin{align*}
\frac{\partial}{\partial \muVec} \FF(\ThetaOLD,\Theta) 
& = \sum\limits_{n=1}^{N} \sum\limits_{\sVec}\int\limits_{\zVec} q_n(\sVec,\zVec;\ThetaOLD)\\
& \phantom{iiiiiiiii} 
 \disT  \Big[- \frac{1}{2}\frac{\partial}{\partial \muVec}
\big(\left( \zVec\, - \,\muVec \right)\odot \sVec\ \big)^\TT\,
\big(\Sigmah \odot \sVec\,\sVec^{\TT}\big)^{-1}\,
\big(\left( \zVec\, - \,\muVec \right)\odot \sVec\ \big) \Big] \dz \\
& = \sum\limits_{n=1}^{N} \sum\limits_{\sVec}\int\limits_{\zVec} q_n(\sVec,\zVec;\ThetaOLD) 
 \disT  \bigg[\big(\Sigmah \odot \sVec\,\sVec^{\TT}\big)^{-1}\,
\Big( \left( \zVec\, - \,\muVec \right) \odot \sVec\ \Big)\bigg] \dz \overset{!}{=} 0 \\ \\
%
%
%
%
\Rightarrow  
\muVec & = \frac{\sum_{n=1}^{N} \E{\sVec\odot\zVec}_n}{\sum_{n=1}^{N}\E{\sVec}_n}.
\end{align*}

\subsection{Optimization of the Latent Covariance}

Lastly we derive the M-step update for the latent covariance $\Sigmah$:
\begin{align*}
\frac{\partial}{\partial \Sigmah}& \FF(\ThetaOLD,\Theta)\\
& = \sum\limits_{n=1}^{N} \sum\limits_{\sVec}\int\limits_{\zVec} q_n(\sVec,\zVec;\ThetaOLD)\\
& \phantom{iiiiiiiii} 
 \disT  \Big[- \frac{1}{2} \frac{\partial}{\partial \Sigmah} \left(\log|\Sigmah \odot \sVec\,\sVec^{\TT}|\right) 
 - \frac{1}{2}\frac{\partial}{\partial \Sigmah}\big(\left( \zVec\, - \,\muVec \right)\odot \sVec\ \big)^\TT\,
\big(\Sigmah \odot \sVec\,\sVec^{\TT}\big)^{-1}\,
\big(\left( \zVec\, - \,\muVec \right)\odot \sVec\ \big) \Big]\dz\displaybreak[1]\\
& = \sum\limits_{n=1}^{N} \sum\limits_{\sVec}\int\limits_{\zVec} q_n(\sVec,\zVec;\ThetaOLD) \\
& \phantom{iiiiiiiii} 
 \disT  \Big[-\frac{1}{2} \big(\Sigmah \odot \sVec\,\sVec^{\TT}\big)^{-1}
 + \frac{1}{2}\big(\Sigmah \odot \sVec\,\sVec^{\TT}\big)^{-2} \big(\left( \zVec\, - \,\muVec \right)\odot \sVec\ \big)\,
\big(\left( \zVec\, - \,\muVec \right)\odot \sVec\ \big)^\TT\, \Big]\dz \overset{!}{=} 0 \\ \displaybreak[1]\\
%
%
%
%
\Rightarrow  
 \Sigmah  & = 
    \sum_{n=1}^{N} \Big[\E{ \left( \zVec\, - \,\muVec \right)\left( \zVec\, - \,\muVec \right)^{\TT}\odot \sVec\,\sVec^{\TT}}_n \Big]
      \odot \Big(\sum_{n=1}^{N}\Big[\E{\sVec\,\sVec^{\TT}}_n \Big]\Big)^{-1} \\
   & = \sum_{n=1}^{N} \Big[\E{(\sVec\odot\zVec)(\sVec\odot\zVec)^{\TT}}_n 
    - \E{\sVec\,\sVec^{\TT}}_n \odot \muVec\muVec^{\TT} \Big] 
    \odot \Big(\sum_{n=1}^{N}\Big[\E{\sVec\,\sVec^{\TT}}_n \Big]\Big)^{-1}.
\end{align*}

\section{Performance vs.\ Complexity Trade-Off}
\label{sec:PerfvsComp}

\begin{figure}
\begin{center}
\includegraphics[width=.95\textwidth]{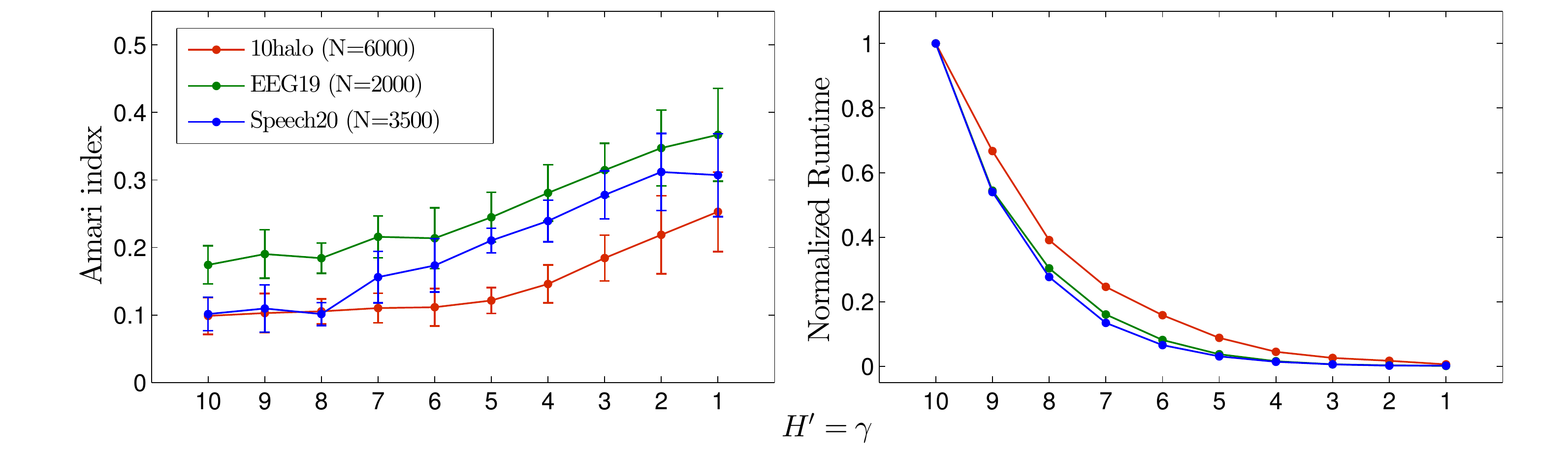}
\caption{Performance of the GSC on 10halo, EEG19 and Speech20 benchmarks for decreasing truncation parameters $H'$ and $\gamma$.
The right plot shows how the computational demand of the truncated variational algorithm decreases with 
decreasing values of the truncation parameters. The runtime plots are normalized by the runtime value obtained for 
$H'=\gamma=10$ for each of the benchmarks.
}
\label{fig:perf-gsc-et}
\end{center}
\end{figure}

\begin{figure}
\begin{center}
\includegraphics[width=.65\textwidth]{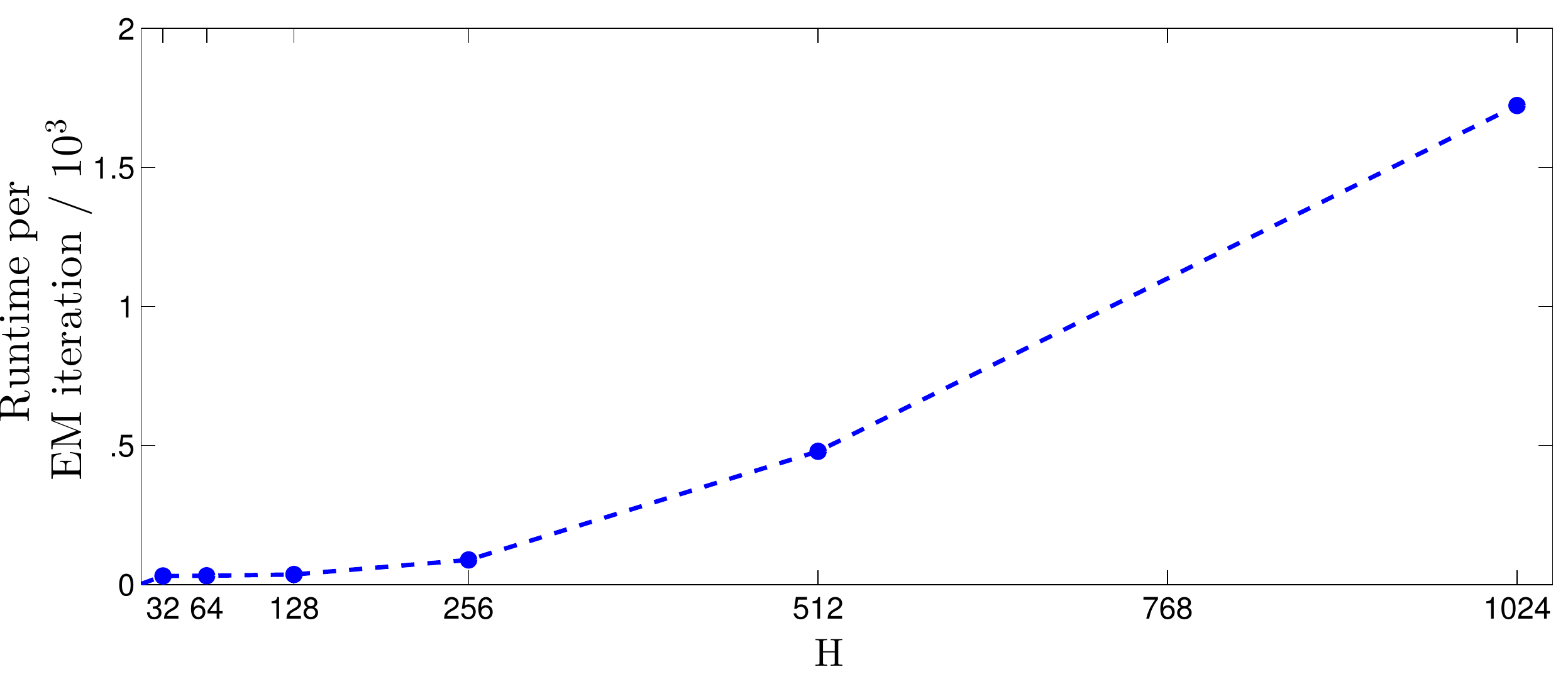}
\caption{Time scaling behavior of GSC for increasing latent dimensions 
$H$ and fixed truncation parameters $H'$ and $\gamma$.}
\label{fig:complexity}
\end{center}
\end{figure}

If the approximation parameters $H'$ and $\gamma$ are held constant, the computational cost
of the algorithm scales with the computational cost of the selection
function. If the latter cost scales linearly with $H$ (as is the case here),
then so does the overall computational complexity \citep[compare complexity
considerations by][]{LuckeEggert2010}). This is consistent with
numerical experiments in which we measured the increase in computational
demand (see Figure \ref{fig:complexity}). In experiments with
$H$ increasing from $16$ to $1024$, we observed a, finally, close to linear increase
of computational costs. However, a larger $H$ implies a larger number of parameters, and thus may
require more data points to prevent over-fitting.  Although a larger
data set increases computational demand, our truncated approximation 
algorithm allows us to take advantage of parallel computing architecture in order to 
more efficiently deal with large data sets (see Appendix \ref{sec:ParallelProcessing} for details). Therefore in practice, we can weaken the extent of an increase in computational cost
due to a higher demand for data.
Furthermore, we examined the benefit of using GSC (in terms of average speedup over EM iterations) versus the cost regarding 
algorithmic performance. We compared approximation parameters in the range of $H' = \gamma = [1,10]$ and again observed the 
performance of the algorithm on the task of source separation (with randomly generated orthogonal ground truth mixing bases and 
no observed noise). Figure \ref{fig:perf-gsc-et} shows that a high accuracy can still be achieved for
relatively small values of $H'$ $\gamma$ which, at the same time, results in strongly reduced
computational demands.

\section{Dynamic Data Repartitioning for Batch/Parallel Processing}
\label{sec:ParallelProcessing}
As described in Section \ref{sec:ET}, the truncated variational approach  
deterministically selects the most likely $H^{\prime}$ causes of a given observation $\yVec$ 
for efficiently approximating the posterior distribution over a truncated latent space. 
In practice one can also use the selected latent causes for applying clustering to 
the observed data, which allows for an efficient and parallelizable batch-mode  
implementation of the E-step of the truncated variational EM algorithm. 

In the batch processing mode, prior to each E-step 
the observed data can be partitioned  
by clustering together the data points w.r.t. their selected latent causes. 
The resulting clusters can then be
processed individually (e.g., on multiple compute cores) to perform the E-step 
(Equations \refp{EqnETPosterior} to \refp{eq:ET-GSC-post}) for all data points in a 
given cluster. 
This approach not only pursues a natural partitioning of data, but in 
a parallel execution environment, it can prove to be more efficient than uniformly 
distributing data \citep[as in][]{BornscheinEtAl2010} among multiple processing units. 
By maximizing the similarity (in latent space) of individual data points 
assigned to each of the processing units, we can overall minimize the number of  
redundant computations involved in Equations \refp{EqnAbbreviations} and \refp{eq:ET-GSC-post}, 
that are tied to specific states of the latents. This can be observed 
by considering Equation \refp{EqnETPosterior}, which is as follows:

\begin{eqnarray}
p(\sVec,\zVec\,|\,\yVecN,\Theta) 
&\approx& \frac{\NGauss(\yVecN;\muVec_{\sVec},C_{\sVec})\,\Bernoulli(\sVec;\piVec)\,\NGauss(\zVec;\,\kappaVecN_{\sVec},\Lambda_{\sVec})}
{\sum_{\sVecPrime\in\KKn}\NGauss(\yVecN;\muVec_{\sVec^{\prime}},C_{\sVec^{\prime}})\,\Bernoulli(\sVec^{\prime};\piVec)\,}\,\delta(\sVec\in\KKn).
\label{eq:ET-GSC-post2}
\end{eqnarray}

\begin{figure}[p]
\centering
\includegraphics[height=3.8cm]{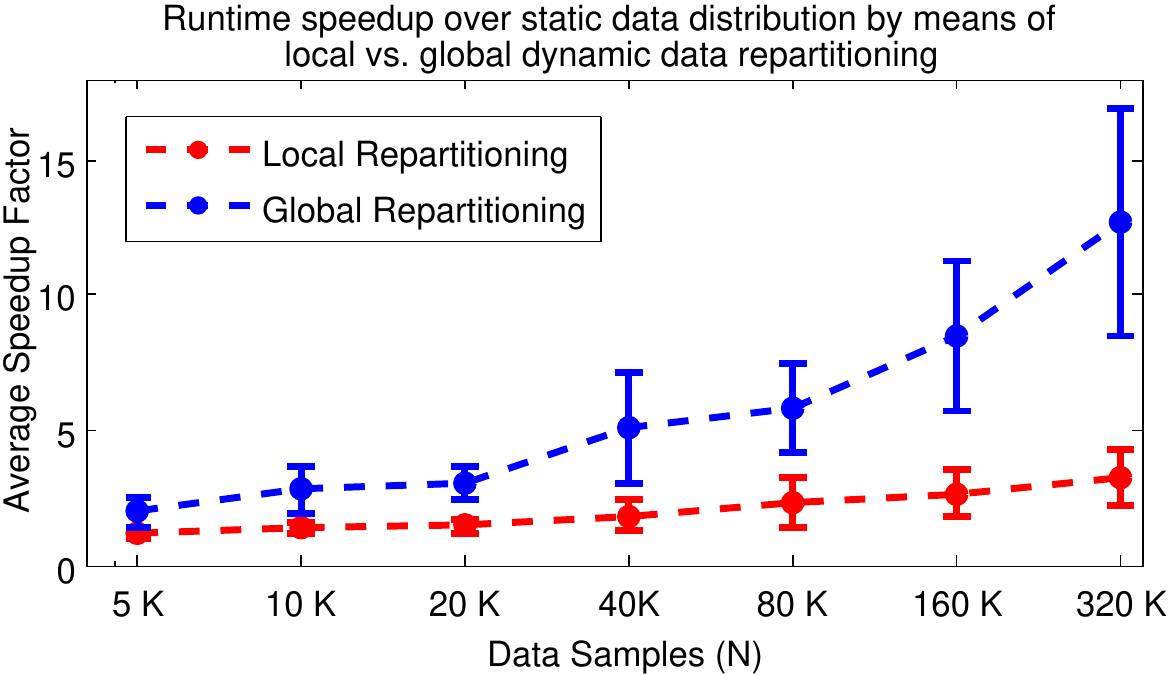}
\caption{Runtime speedup of the truncated variational E-step (Equations \refp{EqnETPosterior} to \refp{eq:ET-GSC-post}) 
with the static data distribution strategy taken as a baseline. The red plot shows the 
speedup when initially uniformly distributed
data samples were only clustered locally by each processing unit, while the blue plot shows the speedup
as a result of globally clustering and redistributing the data. The runtimes include the time taken 
by clustering and repartitioning modules.}
\label{fig:GSC-ET_dyn_rp_spdup}
\end{figure}

Here the parameters $\muVec_{\sVec}, C_{\sVec}$ and $\Lambda_{\sVec}$ entirely depend on 
a particular latent state $\sVec$. Also, $\kappaVecN_{\sVec}$ takes 
prefactors that can be precomputed given the $\sVec$. 
It turns out that to compute \refp{eq:ET-GSC-post2} our clustering-based, dynamic data repartitioning and redistribution strategy 
is more efficient than the uniform data distribution approach of \cite{BornscheinEtAl2010}. This is illustrated in
Figure \ref{fig:GSC-ET_dyn_rp_spdup}, which shows empirical E-step speedup over 
the latter approach taken as a baseline. The
error bars were generated by performing $15$ trials per given data size $N$. For all the 
trials, model scale (i.e., data dimensionality) and truncation approximation parameters were kept constant.\footnote{The 
observed and the latent dimensions of the GSC model 
were $25$ and $20$ respectively. The truncation approximation parameters $H^{\prime}$ and $\gamma$ 
(maximum number of active causes in a given latent state)
were $8$ and $5$ respectively.} Each trial was run in parallel on $24$ computing nodes. 
The red plot in the figure also shows the speedup 
as a result of an intermediate approach. There we initially uniformly distributed the 
data samples which were then only locally clustered by each processing unit at every E-step. The blue plot 
on the other hand shows the speedup as a result of globally clustering and redistributing the
data prior to every E-step. All the reported results here also take into account 
the cost of data clustering and repartitioning.

\begin{figure}[p]
\begin{center}
\includegraphics[width=.85\textwidth]{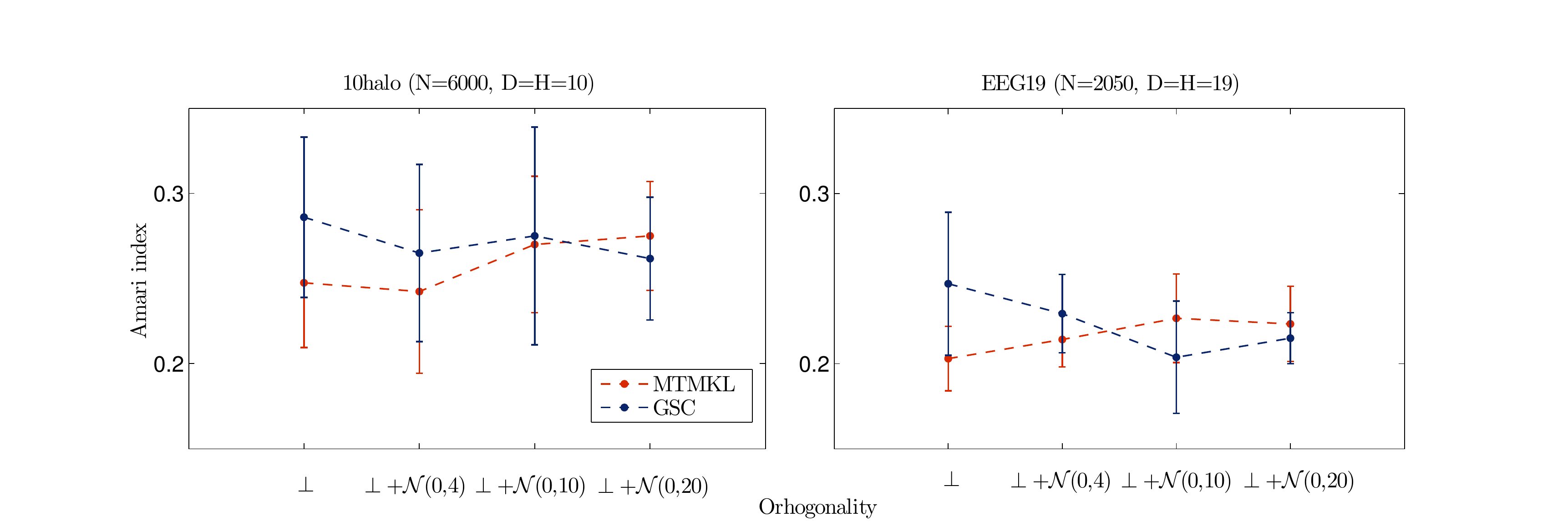}
\caption{Source separation with observation noise. Performance of GSC vs.\ MTMKL
on $\mathrm{10halo}$ and $\mathrm{EEG19}$ benchmarks with varying degrees of orthogonality 
of the mixing bases and Gaussian noise added to observations. Performance of GSC vs.\ MTMKL
on the $\mathrm{Speech20}$ benchmark with varying degrees of orthogonality of the mixing
bases with Gaussian noise added to observed data. The orthogonality on the x-axis 
varies from being orthogonal $\perp$ to increasingly non-orthogonal mixing as 
randomly generated orthogonal bases are perturbed by 
adding Gaussian noise $\NGauss(0,\sigma)$ to them. 
Performance is compared on the Amari index \refp{EqnAmari}. }
\label{fig:gsc_vs_mtmkl_10halo_EEG19_noisy}
\end{center}
\end{figure}
In a parallel setup, we perform the data clustering process by having each processing unit cluster its own 
data locally and then merging the resulting clusters globally. In order to avoid uneven data 
distribution, we also bound the maximum size of a cluster. Currently we pick (per iteration) top
$\alpha$ percentile of occurring cluster sizes as the threshold.\footnote{The 
$\alpha$ for the reported experiments was $5$.} Any cluster larger than
$\alpha$ is evenly broken into smaller clusters of maximum size $\alpha$. Moreover, to minimize communication overhead 
among computational units, we actually only cluster and redistribute the data indices. 
This entails that the actual data must reside in a shared memory structure which is efficiently and 
dynamically accessible by all the computational units. Alternatively, all the units require  
their own copy of the whole data set. 

Here we have introduced and illustrated the gains of dynamic data repartitioning technique in the context of a  
specific sparse coding model, which in fact involves computationally expensive, state-dependent operations for 
computing posterior distributions. The technique however is inherently generic and can be straight-forwardly 
employed for other types of multi-causal models.
\newpage
\bibliographystyle{plainnat}
\bibliography{./gsc}

\begin{thebibliography}{50}
\providecommand{\natexlab}[1]{#1}
\providecommand{\url}[1]{\texttt{#1}}
\expandafter\ifx\csname urlstyle\endcsname\relax
  \providecommand{\doi}[1]{doi: #1}\else
  \providecommand{\doi}{doi: \begingroup \urlstyle{rm}\Url}\fi

\bibitem[Amari et~al.(1995)Amari, Cichocki, and Yang]{AmariEtAl1996}
S.~Amari, A.~Cichocki, and H.~H. Yang.
\newblock A new learning algorithm for blind signal separation.
\newblock In \emph{Advances in Neural Information Processing Systems}, pages
  757--763, 1995.

\bibitem[Attias(1999)]{Attias1999}
H.~Attias.
\newblock Independent factor analysis.
\newblock \emph{Neural Computation}, 11:\penalty0 803--851, 1999.

\bibitem[Bishop(2006)]{Bishop2006}
C.~Bishop.
\newblock \emph{Pattern Recognition and Machine Learning}.
\newblock Springer, 2006.

\bibitem[Bornschein et~al.(2010)Bornschein, Dai, and
  L\"ucke]{BornscheinEtAl2010}
J.~Bornschein, Z.~Dai, and J.~L\"ucke.
\newblock Approximate {EM} learning on large computer clusters.
\newblock In \emph{NIPS Workshop on Big Learning}, 2010.

\bibitem[Bornschein et~al.(2013)Bornschein, Henniges, and
  L\"ucke]{BornscheinEtAl2013}
J.~Bornschein, M.~Henniges, and J.~L\"ucke.
\newblock Are v1 simple cells optimized for visual occlusions? {A} comparative
  study.
\newblock \emph{PLoS Computational Biology}, 9\penalty0 (6):\penalty0 e1003062,
  2013.

\bibitem[Carbonetto and Stephen(2011)]{CarbonettoStephen2011}
P.~Carbonetto and M.~Stephen.
\newblock {Scalable variational inference for Bayesian variable selection in
  regression, and its accuracy in genetic association studies}.
\newblock \emph{Bayesian Analysis Journal}, 2011.

\bibitem[Carvalho et~al.(2008)Carvalho, Chang, Lucas, Nevins, Wang, and
  West]{CarvalhoEtAl2008}
C.~M. Carvalho, J.~Chang, J.~E. Lucas, J.~R. Nevins, Q.~Wang, and M.~West.
\newblock High-dimensional sparse factor modeling: Applications in gene
  expression genomics.
\newblock \emph{Journal of the American Statistical Association}, 103\penalty0
  (484):\penalty0 1438--1456, 2008.

\bibitem[Cichocki et~al.(2007)Cichocki, Amari, Siwek, Tanaka, Phan, and
  Zdunek]{icalab2007}
A.~Cichocki, S.~Amari, K.~Siwek, T.~Tanaka, A.H. Phan, and R.~Zdunek.
\newblock {ICALAB-MATLAB Toolbox Version 3}, 2007.

\bibitem[Dai and L\"{u}cke(2012)]{DaiLucke2012a}
Z.~Dai and J.~L\"{u}cke.
\newblock Autonomous cleaning of corrupted scanned documents --- a generative
  modeling approach.
\newblock In \emph{IEEE Conference on Computer Vision and Pattern Recognition},
  pages 3338--3345, 2012.

\bibitem[Dalen and Gales(2008)]{DalenGales2008}
R.~C. Dalen and M.~J.~F. Gales.
\newblock Covariance modelling for noise-robust speech recognition.
\newblock In \emph{Annual Conference of the International Speech Communication
  Association}, pages 2000--2003, 2008.

\bibitem[Dalen and Gales(2011)]{DalenGales2011}
R.~C. Dalen and M.~J.~F. Gales.
\newblock Extended {VTS} for noise-robust speech recognition.
\newblock \emph{IEEE Transactions on Audio, Speech and Language Processing},
  19\penalty0 (4), 2011.

\bibitem[F\"oldi\'ak(1990)]{Foeldiak1990}
P.~F\"oldi\'ak.
\newblock Forming sparse representations by local anti-{H}ebbian learning.
\newblock \emph{Biological Cybernetics}, 64:\penalty0 165--170, 1990.

\bibitem[Gael et~al.(2008)Gael, Teh, and Ghahramani]{VanEtAl2009}
J.~V. Gael, Y.~W. Teh, and Z.~Ghahramani.
\newblock The infinite factorial hidden {M}arkov model.
\newblock In \emph{Advances in Neural Information Processing Systems}, pages
  1697--1704, 2008.

\bibitem[Garrigues and Olshausen(2007)]{GarriguesOlshausen2007}
P.~Garrigues and B.~A. Olshausen.
\newblock Learning horizontal connections in a sparse coding model of natural
  images.
\newblock In \emph{Advances in Neural Information Processing Systems}, 2007.

\bibitem[Goodfellow et~al.(2013)Goodfellow, Courville, and
  Bengio]{GoodfellowEtAl2013}
I.~Goodfellow, A.~Courville, and Y.~Bengio.
\newblock Scaling up spike-and-slab models for unsupervised feature learning.
\newblock \emph{IEEE Transactions on Pattern Analysis and Machine
  Intelligence}, 35\penalty0 (8):\penalty0 1902--1914, 2013.

\bibitem[Hoyer(2002)]{Hoyer2002}
P.~O. Hoyer.
\newblock Non-negative sparse coding.
\newblock In \emph{Neural Networks for Signal Processing XII: Proceedings of
  the IEEE Workshop on Neural Networks for Signal Processing}, pages 557--565,
  2002.

\bibitem[Ilin and Valpola(2005)]{IlinValpola2005}
A.~Ilin and H.~Valpola.
\newblock On the effect of the form of the posterior approximation in
  variational learning of {ICA} models.
\newblock \emph{Neural Processing Letters}, 22\penalty0 (2):\penalty0 183--204,
  2005.

\bibitem[Jordan et~al.(1999)Jordan, Ghahramani, Jaakkola, and
  Saul]{JordanEtAl1999}
M.~I. Jordan, Z.~Ghahramani, T.~Jaakkola, and L.~K. Saul.
\newblock An introduction to variational methods for graphical models.
\newblock \emph{Machine Learning}, 37\penalty0 (2):\penalty0 183--233, 1999.

\bibitem[Knowles and Ghahramani(2007)]{KnowlesGhahramani2007}
D.~Knowles and Z.~Ghahramani.
\newblock Infinite sparse factor analysis and infinite independent components
  analysis.
\newblock In \emph{Proceedings of International Conference on Independent
  Component Analysis and Signal Separation}, pages 381--388, 2007.

\bibitem[Knowles and Ghahramani(2011)]{KnowlesGhahramani2010}
D.~Knowles and Z.~Ghahramani.
\newblock {Nonparametric Bayesian sparse factor models with application to gene
  expression modeling}.
\newblock \emph{The Annals of Applied Statistics}, 5\penalty0 (2B):\penalty0
  1534--1552, 2011.

\bibitem[Lee et~al.(2007)Lee, Battle, Raina, and Ng]{LeeEtAl2007}
H.~Lee, A.~Battle, R.~Raina, and A.~Ng.
\newblock {Efficient sparse coding algorithms}.
\newblock In \emph{Advances in Neural Information Processing Systems}, pages
  801--08, 2007.

\bibitem[Li and Liu(2009)]{LiLiu2009}
H.~Li and F.~Liu.
\newblock Image denoising via sparse and redundant representations over learned
  dictionaries in wavelet domain.
\newblock In \emph{Proceedings of International Conference on Image and
  Graphics}, pages 754--758, 2009.

\bibitem[L\"ucke and Eggert(2010)]{LuckeEggert2010}
J.~L\"ucke and J.~Eggert.
\newblock Expectation truncation and the benefits of preselection in training
  generative models.
\newblock \emph{Journal of Machine Learning Research}, 11:\penalty0 2855--2900,
  2010.

\bibitem[L\"ucke and Sahani(2008)]{LuckeSahani2008}
J.~L\"ucke and M.~Sahani.
\newblock Maximal causes for non-linear component extraction.
\newblock \emph{Journal of Machine Learning Research}, 9:\penalty0 1227--1267,
  2008.

\bibitem[L\"ucke and Sheikh(2012)]{LuckeSheikh2012}
J.~L\"ucke and A.~S. Sheikh.
\newblock Closed-form {EM} for sparse coding and its application to source
  separation.
\newblock In \emph{Proceedings of International Conference on Latent Variable
  Analysis and Signal Separation}, pages 213--221, 2012.

\bibitem[MacKay(2001)]{MacKay2001}
D.~J.~C. MacKay.
\newblock Local minima, symmetry-breaking, and model pruning in variational
  free energy minimization.
\newblock Online publication: www.inference.phy.cam.ac.uk/mackay/minima.ps.gz,
  2001.

\bibitem[Mairal et~al.(2009)Mairal, Bach, Ponce, and Sapiro]{MairalBPS09}
J.~Mairal, F.~Bach, J.~Ponce, and G.~Sapiro.
\newblock Online dictionary learning for sparse coding.
\newblock In \emph{Proceedings of International Conference on Machine
  Learning}, page~87, 2009.

\bibitem[Mitchell and Beauchamp(1988)]{MitchellBeauchamp1988}
T.~J. Mitchell and J.~J. Beauchamp.
\newblock {Bayesian variable selection in linear regression}.
\newblock \emph{Journal of the American Statistical Association}, 83\penalty0
  (404):\penalty0 1023--1032, 1988.

\bibitem[Mohamed et~al.(2012)Mohamed, Heller, and Ghahramani]{MohamedEtAl2012}
S.~Mohamed, K.~Heller, and Z.~Ghahramani.
\newblock Evaluating {Bayesian and L1} approaches for sparse unsupervised
  learning.
\newblock In \emph{Proceedings of International Conference on Machine
  Learning}, 2012.

\bibitem[Moulines et~al.(1997)Moulines, Cardoso, and Gassiat]{MoulinesEtAl1997}
E.~Moulines, J.-F. Cardoso, and E.~Gassiat.
\newblock Maximum likelihood for blind separation and deconvolution of noisy
  signals using mixture models.
\newblock In \emph{Proceedings of International Conference on Acoustics, Speech
  and Signal Processing}, volume~5, pages 3617--3620, 1997.

\bibitem[Mysore et~al.(2010)Mysore, Smaragdis, and Raj]{GauthamEtAl2010}
G.~J. Mysore, P.~Smaragdis, and B.~Raj.
\newblock Non-negative hidden {M}arkov modeling of audio with application to
  source separation.
\newblock In \emph{Proceedings of International Conference on Latent Variable
  Analysis and Signal Separation}, pages 140--148, 2010.

\bibitem[Neal and Hinton(1998)]{NealHinton1998}
R.~Neal and G.~Hinton.
\newblock A view of the {EM} algorithm that justifies incremental, sparse, and
  other variants.
\newblock In M.~I. Jordan, editor, \emph{Learning in Graphical Models}. Kluwer,
  1998.

\bibitem[Olshausen and Field(1996)]{OlshausenField1996}
B.~Olshausen and D.~Field.
\newblock Emergence of simple-cell receptive field properties by learning a
  sparse code for natural images.
\newblock \emph{Nature}, 381:\penalty0 607--609, 1996.

\bibitem[Olshausen and Field(1997)]{OlshausenField1997}
B.~Olshausen and D.~Field.
\newblock {Sparse coding with an overcomplete basis set: A strategy employed by
  V1?}
\newblock \emph{Vision Research}, 37\penalty0 (23):\penalty0 3311--3325, 1997.

\bibitem[Olshausen and Millman(2000)]{OlshausenMillman2000}
B.~Olshausen and K.~Millman.
\newblock {Learning sparse codes with a mixture-of-Gaussians prior}.
\newblock In \emph{Advances in Neural Information Processing Systems},
  volume~12, pages 841--847, 2000.

\bibitem[Paisley and Carin(2009)]{PaisleyLawrence2009}
J.~Paisley and L.~Carin.
\newblock Nonparametric factor analysis with beta process priors.
\newblock In \emph{Proceedings of International Conference on Machine
  Learning}, pages 777--784, 2009.

\bibitem[Puertas et~al.(2010)Puertas, Bornschein, and
  L\"{u}cke]{PuertasEtAl2010}
G.~Puertas, J.~Bornschein, and J.~L\"{u}cke.
\newblock The maximal causes of natural scenes are edge filters.
\newblock In \emph{Advances in Neural Information Processing Systems},
  volume~23, pages 1939--47, 2010.

\bibitem[Ranzato and Hinton(2010)]{RanzatoHinton2010}
M.'A. Ranzato and G.~Hinton.
\newblock Modeling pixel means and covariances using factorized third-order
  boltzmann machines.
\newblock In \emph{IEEE Conference on Computer Vision and Pattern Recognition},
  pages 2551--2558, 2010.

\bibitem[Rattray et~al.(2009)Rattray, Stegle, Sharp, and Winn]{RattrayEtAl2009}
M.~Rattray, O.~Stegle, K.~Sharp, and J.~Winn.
\newblock Inference algorithms and learning theory for {B}ayesian sparse factor
  analysis.
\newblock \emph{Journal of Physics: Conference Series}, 197:\penalty0 012002
  (10pp), 2009.

\bibitem[Ribeiro and Opper(2011)]{RibeiroOpper2011}
F.~Ribeiro and M.~Opper.
\newblock Expectation propagation with factorizing distributions: A gaussian
  approximation and performance results for simple models.
\newblock \emph{Neural Computation}, 23:\penalty0 1047--1069, 2011.

\bibitem[Seeger(2008)]{Seeger2008}
M.~Seeger.
\newblock Bayesian inference and optimal design for the sparse linear model.
\newblock \emph{Journal of Machine Learning Research}, 9:\penalty0 759--813,
  2008.

\bibitem[Shelton et~al.(2011)Shelton, Bornschein, Sheikh, Berkes, and
  L\"ucke]{SheltonEtAl2011}
J.~Shelton, J.~Bornschein, A.-S. Sheikh, P.~Berkes, and J.~L\"ucke.
\newblock Select and sample --- a model of efficient neural inference and
  learning.
\newblock In \emph{Advances in Neural Information Processing Systems}, pages
  2618--2626, 2011.

\bibitem[Spratling(2006)]{Spratling2006}
M.~Spratling.
\newblock Learning image components for object recognition.
\newblock \emph{Journal of Machine Learning Research}, 7:\penalty0 793--815,
  2006.

\bibitem[Suzuki and Sugiyama(2011)]{SuzukiSugiyama2011}
T.~Suzuki and M.~Sugiyama.
\newblock Least-squares independent component analysis.
\newblock \emph{Neural Computation}, 23\penalty0 (1):\penalty0 284--301, 2011.

\bibitem[Teh et~al.(2007)Teh, G\"or\"ur, and Ghahramani]{TehGorurGhar2007}
Y.~W. Teh, D.~G\"or\"ur, and Z.~Ghahramani.
\newblock {Stick-breaking construction for the Indian buffet process.}
\newblock \emph{Journal of Machine Learning Research}, 2:\penalty0 556--563,
  2007.

\bibitem[Titsias and Lazaro-Gredilla(2011)]{TitsiasLazaro2011}
M.~Titsias and M.~Lazaro-Gredilla.
\newblock Spike and slab variational inference for multi-task and multiple
  kernel learning.
\newblock In \emph{Advances in Neural Information Processing Systems}, pages
  2339--2347, 2011.

\bibitem[Turner and Sahani(2011)]{TurnerSahani2011}
R.~E. Turner and M.~Sahani.
\newblock {Two problems with variational expectation maximisation for
  time-series models}.
\newblock In \emph{Bayesian Time Series Models}. Cambridge University Press,
  2011.

\bibitem[West(2003)]{West2003}
M.~West.
\newblock Bayesian factor regression models in the ``large p, small n''
  paradigm.
\newblock \emph{Bayesian Statistics}, pages 723--732, 2003.

\bibitem[Yoshida and West(2010)]{YoshidaEtAl2010}
R.~Yoshida and M.~West.
\newblock Bayesian learning in sparse graphical factor models via variational
  mean-field annealing.
\newblock \emph{Journal of the American Statistical Association}, 99:\penalty0
  1771--1798, 2010.

\bibitem[Zhou et~al.(2009)Zhou, Chen, Paisley, Ren, Sapiro, and
  Carin]{ZhouetAl2009}
M.~Zhou, H.~Chen, J.~Paisley, L.~Ren, G.~Sapiro, and L.~Carin.
\newblock Non-parametric {B}ayesian dictionary learning for sparse image
  representations.
\newblock In \emph{Advances in Neural Information Processing Systems}, pages
  2295--2303, 2009.

\end{thebibliography}
\end{document}